\documentclass[twocol]{ametsocV6.1}
\usepackage{url}

\title{A Machine Learning Tutorial for Operational Meteorology, Part II: Neural Networks and Deep Learning}
\authors{Randy J. Chase\correspondingauthor{Randy J. Chase, randy.chase@colostate.edu}\aff{a,b,c}, David R. Harrison\aff{b,d,e}, Gary M. Lackmann\aff{f} and Amy McGovern\aff{a,b,c}} 

\affiliation{\aff{a}{School of Computer Science, University of Oklahoma, Norman OK USA}\\
\aff{b}{School of Meteorology, University of Oklahoma, Norman OK USA}\\
\aff{c}{NSF AI Institute for Research on Trustworthy AI in Weather, Climate, and Coastal Oceanography, University of Oklahoma, Norman OK USA}\\
\aff{d}{Cooperative Institute for Severe and High-Impact Weather Research and Operations, University of Oklahoma, Norman OK USA}\\
\aff{e}{NOAA/NWS/Storm Prediction Center, Norman, Oklahoma}\\
\aff{f}{Department of Marine, Earth, and Atmospheric Sciences, North Carolina State University, Raleigh, North Carolina}\\}

\abstract{Over the past decade the use of machine learning in meteorology has grown rapidly. Specifically neural networks and deep learning have been used at an unprecedented rate. In order to fill the dearth of resources covering neural networks with a meteorological lens, this paper discusses machine learning methods in a plain language format that is targeted for the operational meteorological community. This is the second paper in a pair that aim to serve as a machine learning resource for meteorologists. While the first paper focused on traditional machine learning methods (e.g., random forest), here a broad spectrum of neural networks and deep learning methods are discussed. Specifically this paper covers perceptrons, artificial neural networks, convolutional neural networks and U-networks. Like the part 1 paper, this manuscript discusses the terms associated with neural networks and their training. Then the manuscript provides some intuition behind every method and concludes by showing each method used in a meteorological example of diagnosing thunderstorms from satellite images (e.g., lightning flashes). This paper is accompanied with an open-source code repository to allow readers to explore neural networks using either the dataset provided (which is used in the paper) or as a template for alternate datasets.}

\begin{document}

\maketitle

%
%
%
%
%
%

%

\section{Introduction}
In the previous part of this tutorial series \citet{Chase2022} (hereafter Part 1) provided a survey of many of the most common traditional machine learning techniques that a meteorologist might encounter. This included: linear regression, logistic regression, naive bayes, decision trees, random forest, gradient boosted trees and support vector machines. Beyond discussing the formulation of the methods, Part 1 also discussed the general terms associated with machine learning and provided an end-to-end machine learning example to detect lightning flashes within satellite and radar images. In this manuscript we continue our explanation and tutorial of supervised machine learning techniques by discussing a rapidly expanding category of machine learning known as \textit{neural networks} and \textit{deep learning}.

While neural networks can be viewed similarly to the other methods described in Part 1 (i.e., an empirical tool for making predictions and classifications), there are numerous nuances and different terms associated with neural networks that motivate their own detailed discussion. Furthermore, given the accelerated growth of neural networks (c.f., Fig. 1e in Part 1) and recent impressive demonstrations of neural networks achieving similar forecasting performance to numerical weather prediction \citep[e.g.,][]{Weyn2020,Rasp2021,Ravuri2021,Espeholt2022,Keisler2022,Lam2022,Bi2022,Nguyen2023}, the meteorological literature could benefit from a neural-network specific \textit{plain language} discussion and simple meteorological example. 

This paper follows the same organization as Part 1. Section 2 provides an introduction to neural-network based machine learning methods and defines common neural network terms. Section 3 discusses how the neural network methods discussed in Section 2 can be applied to a meteorological example. Section 4 summarizes this paper. The specific neural network types covered in this manuscript are perceptrons, artificial neural networks, convolutions neural networks and "U" shaped networks (U-Net). 

\section{Neural network methods and common terms\label{sec:MethodsTerms}}
This section introduces many of the common terms that meteorologists would encounter while reading about or using output from neural networks. The goal of this paper is to provide readers with the intuition behind the different neural network methods as well as introduce common terms used within neural networks so that readers can become familiar with them. This section will dive deeper than Part 1's corresponding section in order to remove some of the mystery of the more complex mathematical nature of neural networks and hopefully achieve the same level of intuition as the traditional methods. 

Before describing the various types of neural networks, also known as different \textit{architectures}, we first define neural networks as: the group of machine learning methods that use a network of trainable weights that are organized in a structure that loosely resemble a biological brain. The name \textit{neural network} comes from the analogy of how the information is passed in a biological brain and more specifically across neurons. Simply, a biological brain observes some information which is then processed by a neuron and passed along a series of connections to numerous other neurons resulting in a thought or action.

Another common term that is used with neural networks is \textit{deep learning}. While deep learning is often perceived as a synonym of neural networks by new users, it is actually a specific subset of neural networks. Since there are many different definitions of what exactly deep learning is, deep learning is defined here as a neural network that contains a minimum of two or more \textit{hidden layers}\footnote{hidden layers are layers that don’t directly interact with the input or output of a neural network. These are discussed more later}, though often involves many more than two layers (e.g., 10s to 100s). This deep learning definition can be interpreted as a minimum complexity requirement for a neural network to be considered deep learning.

\subsection{Architectures of neural networks \label{sec:arch}}
\subsubsection{The perceptron}

\begin{figure*}[t]
 \centering
 \noindent\includegraphics[width=5in]{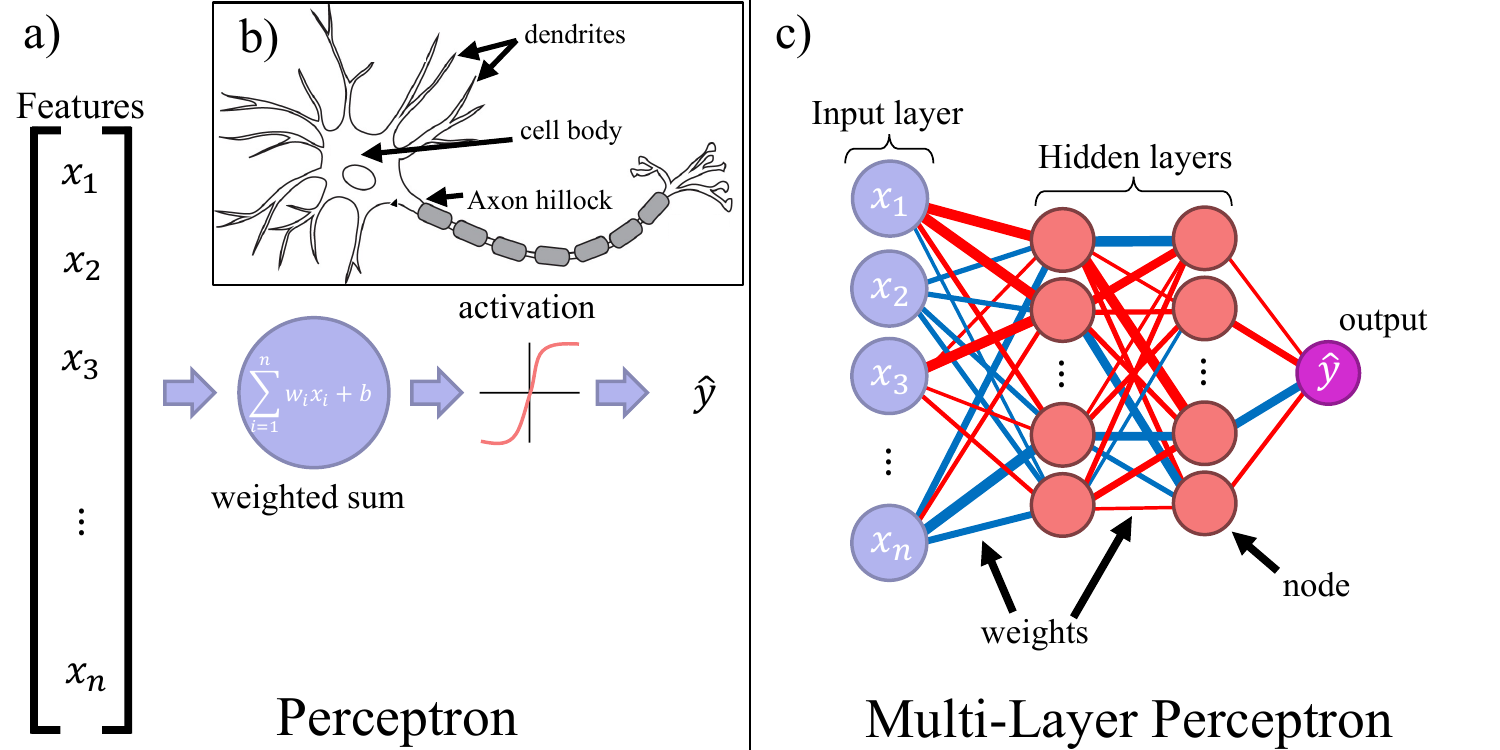}\\
 \caption{Schematic of (a) a perceptron (b) a biological neuron adapted from \citet{Henley2021} and (c) a multi-layer perceptron.} \label{perceptron_mlp}
\end{figure*}

The first architecture of neural networks came from \citet{McCulloch1943} in their formulation of a \textit{logical neuron}, called later a \textit{perceptron} (and referred to later in this document as nodes). A schematic of a perceptron is in Fig. \ref{perceptron_mlp}a. The perceptron has some input data (i.e., features), which are altered by weights and aggregated (i.e., summed). Then the aggregated value is passed through an \textit{activation function} which determines the output of the perceptron. This is similar to a biological neuron (Fig. \ref{perceptron_mlp}b; \citet{Henley2021}), where information is passed to the neuron from the dendrites, aggregated at the cell body, passed through the axon hillock function and then results in some output of the neuron. 

Mathematically, the perceptron is
\begin{equation}
    f(x) = \sigma (\sum_{i=1}^{i=n} w_i x_i + b), \label{e1}
\end{equation}
where $w_i$ are the weights, $x_i$ are the $n$ total input features, $b$ is the bias and $\sigma$ is the activation function. Equation 1 will look familiar to those who read Part 1 because in essence it is the same as linear and logistic regression (Equation 1 in Part 1). In fact, Eq. \ref{e1} is exactly logistic regression if the activation function is the sigmoid function. The only difference is how the weights, $w$, are determined which is discussed later (Section 2.b). Since it is effectively the same as logistic regression, the perceptron is used in a similar manner. For example, we could use the same input features as Part 1 (e.g., minimum brightness temperature) to determine if there were any lightning flashes in a satellite image. Given the limited representational capacity of a perceptron, their application in the meteorological literature has been limited. One meteorological example can be found in \citet{Kim2013}, where a perceptron is used to remove chaff\footnote{military aircraft countermeasure for heat-seeking missiles} and clutter from radar data. 

\subsubsection{Multi-layer perceptron (Artificial Neural Network; ANN) }

\begin{figure*}[t]
 \centering
 \noindent\includegraphics[width=5.5in]{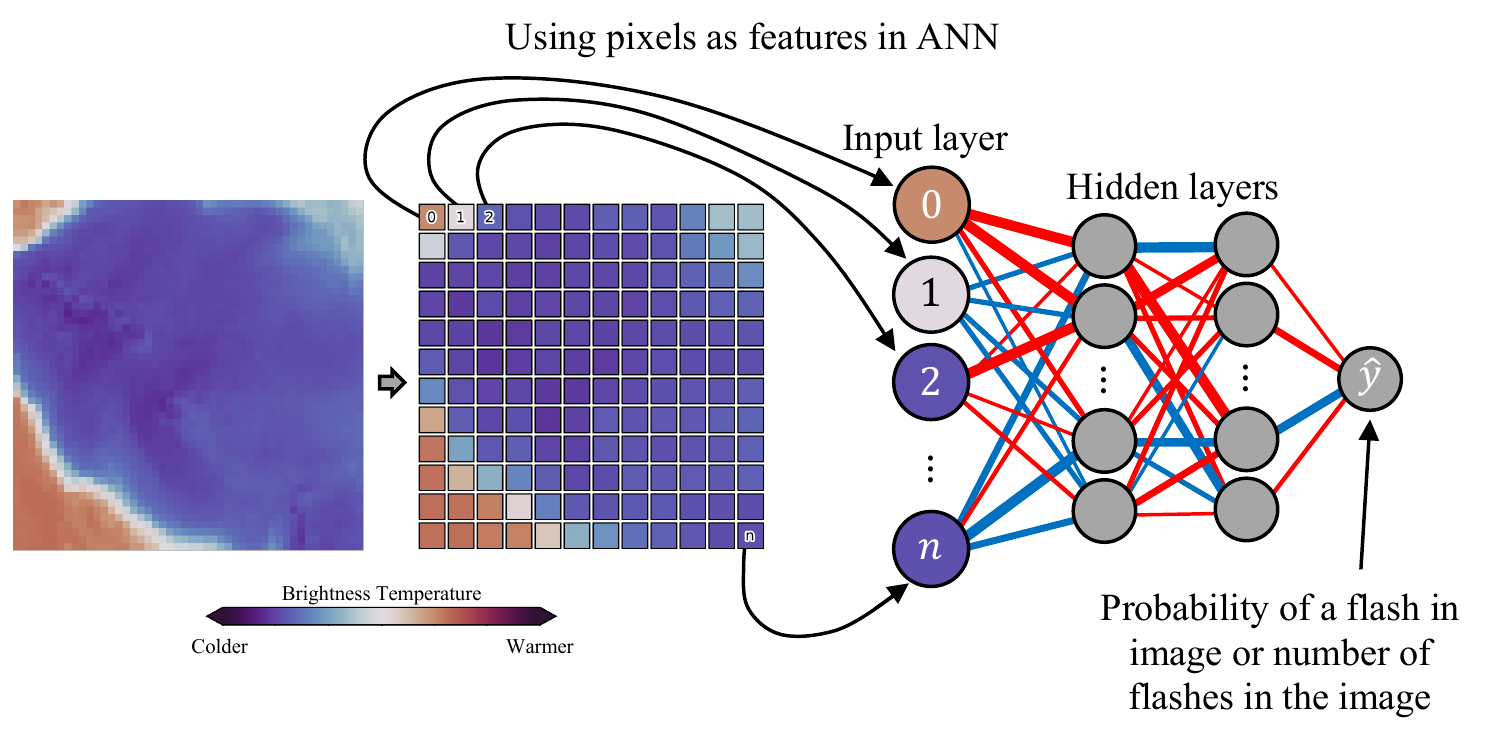}\\
 \caption{Schematic of using each image pixel as input features to a multi-layer perceptron (also known as an Artificial Neural Network [ANN]). The leftmost image is the infrared brightness temperature. The second image is the same brightness temperature image is the same image but coarsened for visualization purposes.}\label{pixel_input}
\end{figure*}

Akin to how many neurons make up a brain, the second type of neural network is an extension of the single perceptron which includes multiple perceptrons and multiple layers of multiple perceptrons \citep{Rumelhart1986}. This \textit{multi-layer perceptron} network is also known as an \textit{Artificial Neural Network} (ANN; Fig. \ref{perceptron_mlp}b). Similar to the single perceptron, the data flows from the input layer (i.e., input data) to each of the perceptrons (hereafter \textit{nodes}) through an activation function. The resulting information is then passed to all of the nodes in the next layer and so on until it reaches the output layer (i.e., where the final prediction is made). Any layer of nodes that are between the input and output are known as \textit{hidden layers}. Mathematically, the multi-layer network is usually summarized by the following
\begin{equation}
    \hat{y} = f( \textbf{x} ; \theta) \label{e2}
\end{equation}
where $\hat{y}$ is the output of the neural network, $f$ is the neural network which is a function of the input data $\textbf{x}$ and has parameters (i.e., weights and biases) $\theta$. Like the perceptron, the same features from the Part 1 data example can be plugged in as the input layer (Fig. \ref{perceptron_mlp}b). Alternatively, ANNs can efficiently handle images where each pixel can be used as a feature (Fig. \ref{pixel_input}). Both methods are shown in the meteorology example in Section 3.

Use of ANNs has been much more popular in meteorology than a single perceptron. Initial applications of ANNs in meteorology date back to the 1990s, which included short-term forecasts of: rain \citep{Kuligowski1998}; road temperatures \citep{Shao1998}; significant thunderstorms \citep{McCann1992}; damaging winds \citep{Marzban1998} and even tornadoes \citep{Marzban1996}. More recent examples include: short-term forecasting of solar irradiance \citep{McCandless2016}; building radar retrievals of snowfall \citep{Chase2021}; and forecasting tropical cyclone intensity \citep{Cloud2019,Xu2021}.

Before continuing to the next type of neural networks, a popular neural network-based tool should be mentioned: \textit{Self-Organizing Maps} \citep[SOM;][]{Kohonen1997}. Self-organizing maps are neural networks but they are an \textit{unsupervised} machine learning method (recall the discussion of supervised and unsupervised machine learning in Part 1). Thus, their task is often to \textit{cluster} data without human prescribed classes. For example, SOM have been used to classify severe storm environments \citep{Anderson-Frey2017,Katona2021}, organize synoptic weather patterns in context of warm precipitation events \citep{Wang2019}, and auto classify near-proximity soundings to supercells \citep{Nowotarski2013}. While these unsupervised clustering applications are useful, they are not the focus of these two manuscripts (Part 1 and Part 2) and likely deserve to have their own dedicated manuscript discussing all unsupervised techniques (e.g., Principle Component Analysis, K-means clustering).

\subsubsection{Convolutional neural network (CNN)}

\begin{figure*}[t]
 \centering
 \noindent\includegraphics[width=6in]{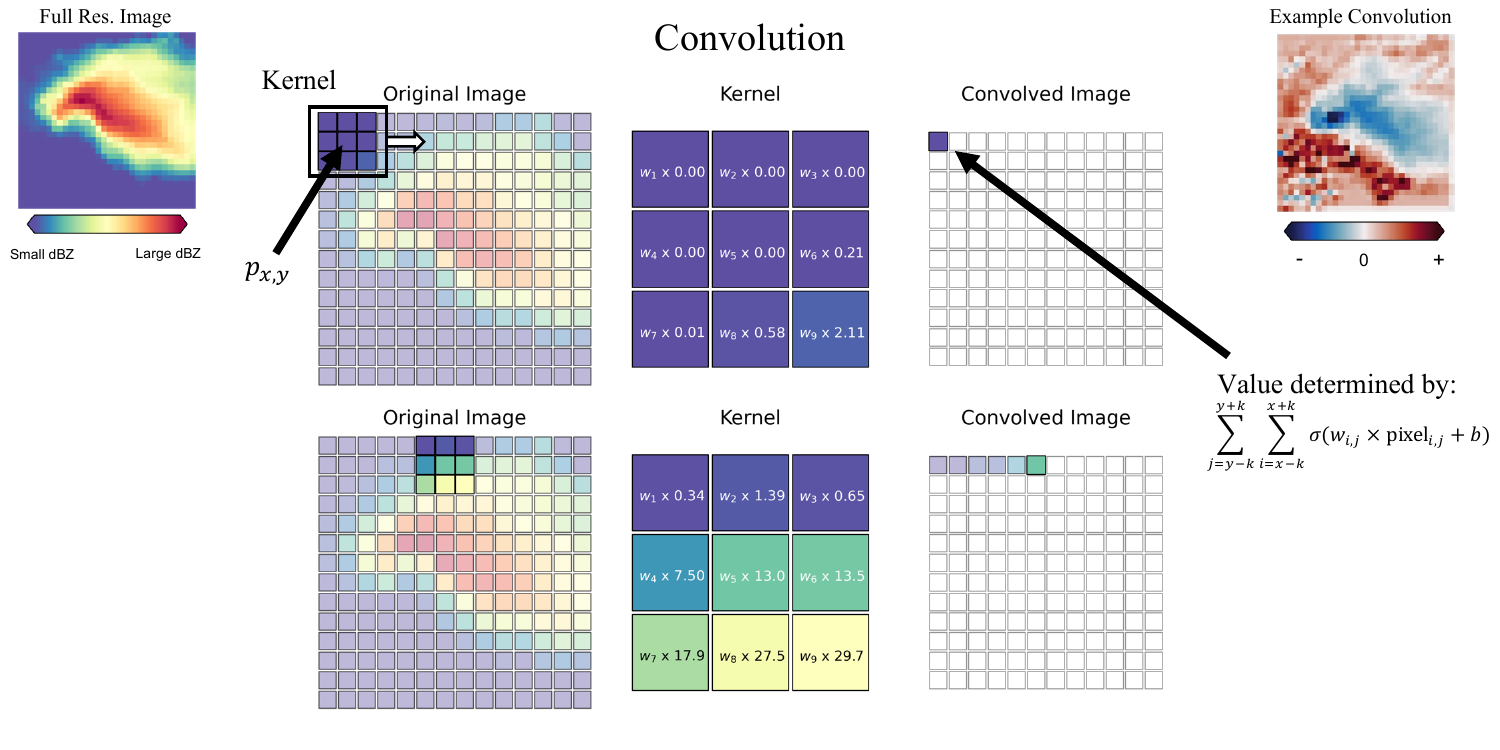}\\
 \caption{Convolution graphic. The original hook echo radar reflectivity is located in the top left corner. The convolution at step 0 is shown in the top row while the convolution at step 6 is shown in the bottom row. Note these images are coarsened for visualization purposes (i.e., can see the pixels). An animation of this convolution can be found in the Notebook 6 in the code repository. The result of the full convolution is shown in the top right, blues are negative, reds are positive and they are unitless.}\label{convolution}
\end{figure*}

While applications of ANNs can be impressive, an additional advancement to neural networks was introduced by \citet{Lecun1989} named \textit{Convolutional Neural Networks} (CNN). As the name implies, these neural networks use \textit{convolutions} where a convolution is a function that processes an image by systematically altering the image with a small window called a \textit{kernel} or \textit{filter}. Graphically a convolution is shown in Fig. \ref{convolution}. An image is convolved/filtered by moving this kernel through the image. The kernel is made up of weights (center of Fig. \ref{convolution}), much like the nodes in an ANN, which are used to create a weighted sum that is a convolved image (also known as a feature map). Note that the weighted sum is passed through an activation function, as was done in the ANN. The mathematical expression of a convolution for some pixel $p$ with coordinates $x,y$ ($p_{x,y})$ is:
\begin{equation}
    p_{x,y}= \sigma(\sum_{j=y-k}^{j=y+k}\sum_{i=x-k}^{i=x+k} w_{i,j} p_{i,j} + b), \label{e3}
\end{equation}
where $w_{i,j}$ is a scalar value (i.e., weight) that is learning during training at the at the $i^{th}$ and $j^{th}$ coordinate, $p_{i,j}$ is the pixel value at the same $i^{th}$ and $j^{th}$ coordinate, $k$ is the floor (i.e., rounded down) of half the kernel size\footnote{for the example in Fig. \ref{convolution} the kernel size is 3}, $b$ is a scalar constant (also known as a bias term) and $\sigma$ is an activation function (e.g., see Sigmoid in Part 1). This equation is then repeated for all pixels in the image. For visual learners, we encourage readers to check out the animated images in \citet{Lagerquist2020b}'s supplemental material\footnote{\url{https://journals.ametsoc.org/view/journals/mwre/148/7/mwrD190372.xml?tab_body=supplementary-materials}} as well as Notebook 6 in the accompanying code with this manuscript\footnote{\url{https://github.com/ai2es/WAF_ML_Tutorial_Part2/blob/main/jupyter_notebooks/Notebook6_Convolutions.ipynb }}. You might notice that the convolution equation doesn't work for the edge of an image (i.e., negative indices don't make sense in this context). The fix for the edges of the image is to \textit{pad} (i.e., add) a row of zeros on all edges of the image. 

\begin{figure*}[t]
 \centering
 \noindent\includegraphics[width=6in]{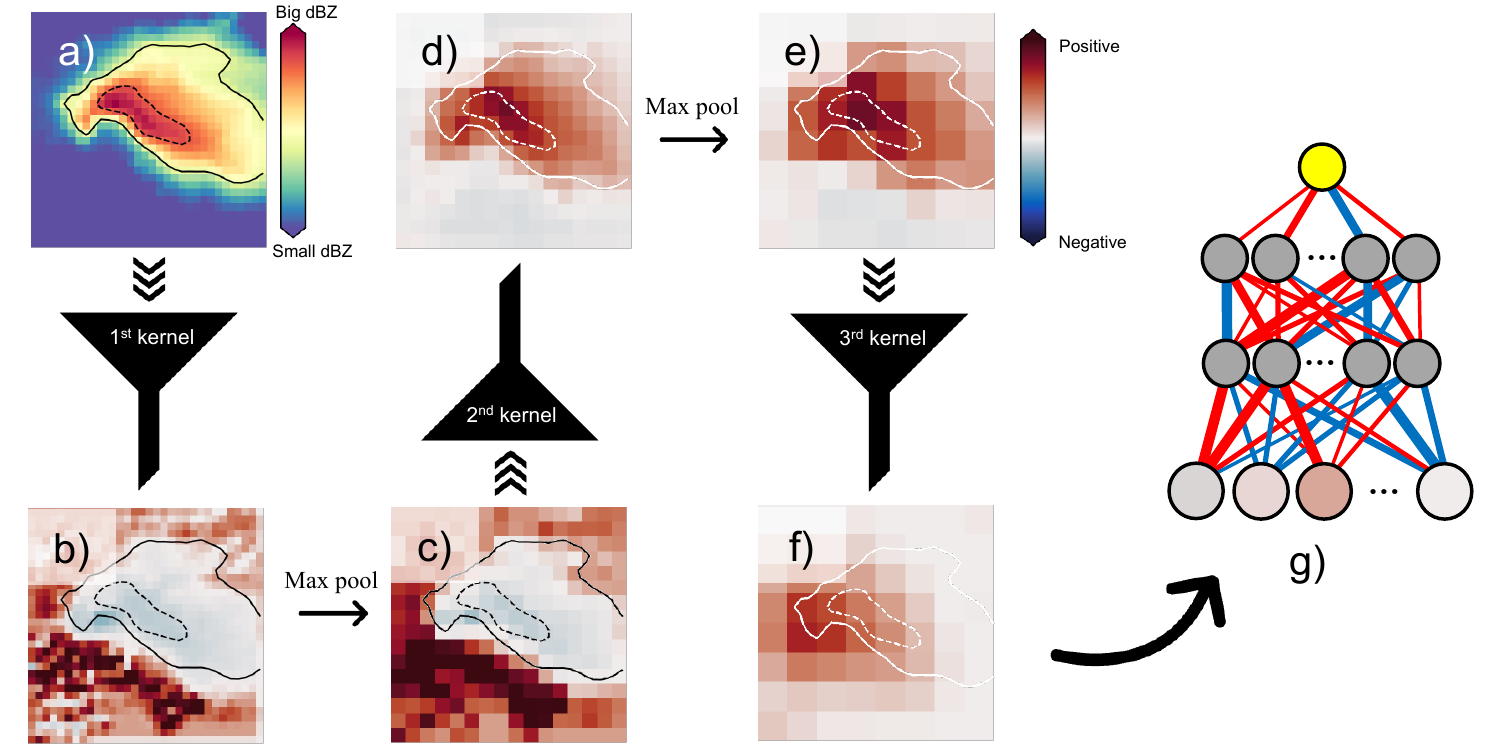}\\
 \caption{A schematic showing how the learned kernels/filters from a CNN extracts features. (a) The same hook echo example from Fig. \ref{convolution}, but the 25 and 50 dBZ contours are drawn. Colors are the normalized reflectivity values (b) the output of the first convolution, reds are positive, blues are negative (unitless). The same storm contours are included from a. (c) the result of pooling the image in b. (d) same as b, but taking the convolution of c. (e) result of pooling d. (f) the same as d, but taking the convolution of e (g) an ANN that takes the pixels of f as input.}\label{convolution_multi}
\end{figure*}

The idea of a convolution is probably very abstract, so let's consider an in depth example of how one could work. Figure \ref{convolution_multi}a shows a classic radar 'hook' echo \citep{Fujita1953}. The data are from \citet{Lagerquist2020b} where the goal is to determine if the storm in the radar image will produce a tornado in the next hour. Before jumping into the CNN, first consider how a human would extract information from a radar image that might be useful for determining if a tornado will occur. One thought could be that we could have meteorologists go through thousands of images and encode 'hook' (i.e., 0 for no hook, 1 for hook), but that would be labor intensive and subjective. Another thought would be to take the max reflectivity of this image. That could work, since stronger storms have stronger updrafts and stronger reflectivity, which could be more likely to create a tornado, but maximum reflectivity is likely too simple. This thought activity should have illustrated that the optimal choices of data to extract are not trivial and since a machine learning model can only be as good as the predictors it is given, determining skillful inputs (i.e., features) is vital. 

One of the main benefits of a CNN is that it will extract relevant features (i.e., patterns in input data) automatically from the data it is provided in order to optimize performance. Thus, there is no need for a human to manually identify important patterns in the images. Furthermore, since the CNN is using these convolutional windows, spatial information is automatically encoded into the features. The CNN does this feature extraction through the learning of the weights of the kernels. Sometimes these kernels are referred to as \textit{filters} which is likely a more apt description of them. The kernels \textit{filter} the features from the image. How the specific weights are learned is discussed in the following section (Section 2.b), but know that the CNN tries multiple filters which result in some amount of error (e.g., truth - ML prediction). This error is then used to inform the CNN which filters work better than others and how to tweak the filters to get better performance (i.e., less error).

While the auto-extraction of relevant features is a benefit of CNNs, it can also lead to unexpected results. A non-meteorology example is from \citet{Lapuschkin2019} where the machine learning task it the classification of images with classes of dog, cat, horse etc. \citet{Lapuschkin2019} showed that the CNN was using the copyright of images as a dominate predictor of the horse images. The \citet{Lapuschkin2019} example illustrates how vital the interrogation of the decisions of a CNN, and more broadly all machine learning methods, is. The interrogation of machine learning methods include eXplainable Artificial Intelligence (XAI) methods which are discussed in Section 3.f. 

Back to the hook echo example (Fig. \ref{convolution_multi}a). One of the learned filters is shown in Fig. \ref{convolution_multi}b, which appears to filtering out the storm location. But notice, that after a single convolution, we are still stuck with the same scenario from before: how do we extract information from the new image? (Fig. \ref{convolution_multi}b). To answer this question, several more convolutions and many filters are typically used with an additional layer, called a \textit{pooling layer}, in between convolution layers. A pooling layer is a way of reducing the dimensionality of the image, which ultimately allows the CNN to distill high resolution information into useful features. One can view pooling as making an image a lower resolution, like converting high resolution precipitation maps from one kilometer horizontal grid spacing to a more regional scale such as 20 km. The intuition behind pooling layers can be thought of as summarizing the key findings of a scientific paper. The pooling layers boil down the most vital information in the paper (image), representing it in a smaller space (less pixels). Pooling is done similarly to the convolution kernel (i.e., uses a window), but has static weights which either take an average value (i.e., average pooling) or passes the maximum value through (i.e., maximum pooling). The typical size of a pooling kernel is two by two, which effectively halves the dimensions of the image. In the hook echo example the result of the pooling is apparent as the grid becomes coarser and the CNN focuses (i.e., large values) in on the hook echo location (Fig. \ref{convolution_multi}f). 

In the process of summarizing data (i.e., pooling), there is less space for information to be stored (i.e., less pixels). Thus, in the CNN the number of filters (i.e., kernels/filters) typically increases with depth in the CNN (Figure 5). Drawing on the same scientific paper analogy, imagine at the beginning of the network the CNN has only one filter and it writes a full-page summary on one key finding of the paper with plenty of detail. After another pooling layer, the full-page summary gets summarized further into one paragraph. Another pooling layer results in a sentence, and finally another layer leaves the filter with one word. The analogy here keeps the number of filters as the same (i.e., one). If instead the CNN has access to more filters as it goes deeper (i.e., more pooling layers), the CNN can then write summarize different aspects of the key findings, enhancing the total extracted information by the CNN.

\begin{figure*}[t]
 \centering
 \noindent\includegraphics[width=6in]{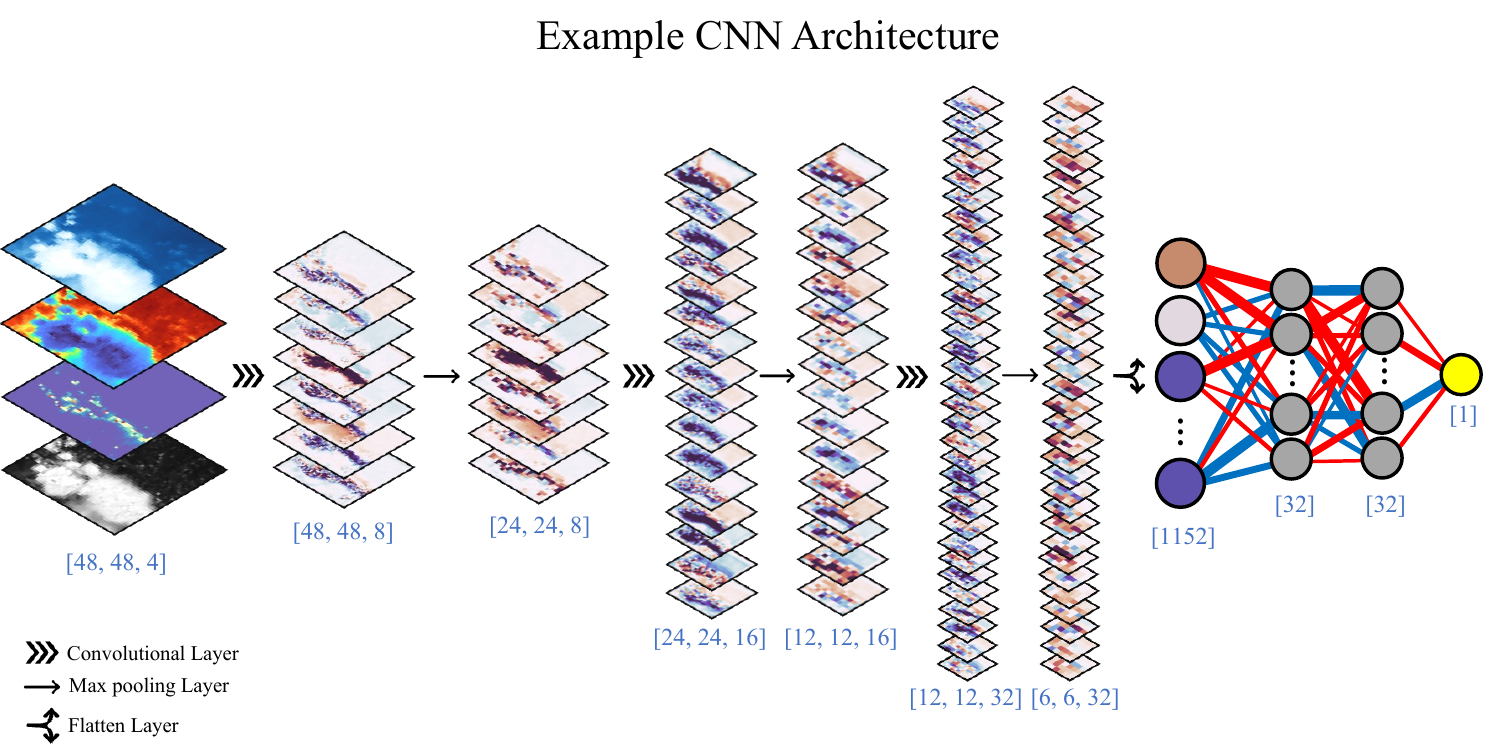}\\
 \caption{Example convolutional neural network (CNN) architecture. The different arrow indicators symbolize different layer types, see the legend in the lower left hand corner. The blue bracketed text is the size of the images ([x-dimension, y-dimension, channel/feature dimension]) or vector (i.e., dense layers). This is the exact architecture for the best performing CNN in Section 3.}\label{convolution_arch}
\end{figure*}

After some number of convolutional layers and their corresponding pooling layers (the exact number determined by hyperparameter tuning), usually an ANN is appended on to the end (Fig. \ref{convolution_multi}g). In other words, after the final convolutional layer, the images are reshaped into a one dimensional vector (like Fig. \ref{pixel_input}) and passed into the ANN. An example of a CNN architecture that is used in the data example (Section 3.d) is shown in Fig. \ref{convolution_arch}. 

Convolutional neural networks are an emerging technique in the meteorological literature that can do complex tasks. Examples include: detecting fronts in reanalysis data \citep{Lagerquist2019,Lagerquist2020a}; estimating tropical cyclone intensity for satellite data \citep{Chen2019,Griffin2022}; determining if a storm will produce severe hail \citep{Gagne2019}; automatically classifying strongly rotating storms in numerical weather prediction data \citep{Molina2021} and identifying intense convection on satellite imagery \citep{Cintineo2020}. While all of the discussion thus far has been focused on two dimensional convolutions and images, the idea can be extended to work on one dimensional data (e.g., a temperature profile) and to full three dimensional volumes (e.g., numerical weather prediction output) or as time as a third dimension. The only change to go from a two dimensional convolution to a one or three dimensional convolution is the shape of the kernel. Both one-dimensional \citep[e.g.,][]{Stock2021,Harrison2022} and three dimensional convolutions have been used in meteorological applications \citep[e.g.,][]{Lagerquist2020b,Zhou2020,Kamangir2021,Justin2022}.

\subsubsection{"U" Network (U-Net)}

\begin{figure*}[t]
 \centering
 \noindent\includegraphics[width=6in]{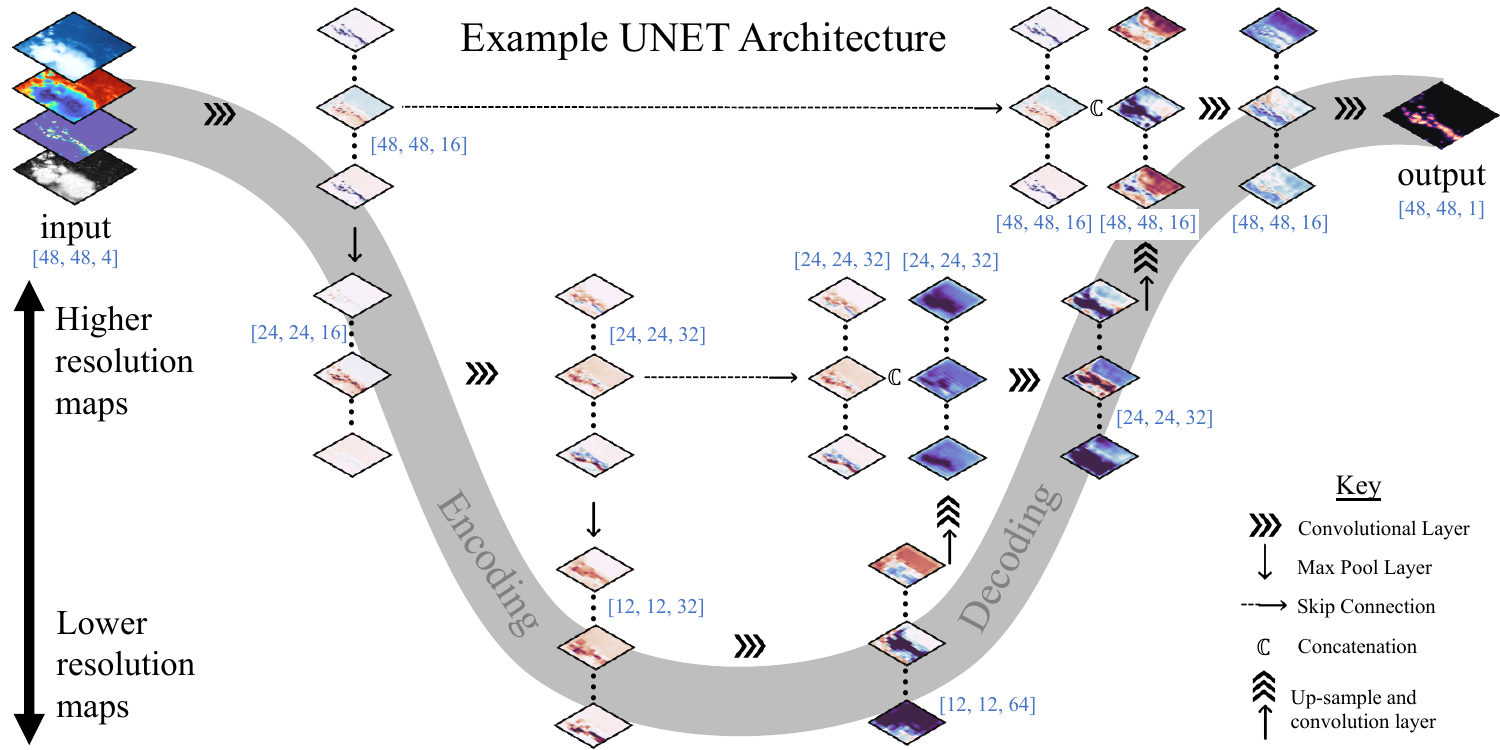}\\
 \caption{Example U-Net architecture. Like Fig. \ref{convolution_arch}, the different arrow indicators symbolize different layer types and the legend is in the lower right hand corner. The red-blue colored images in the middle are the convolved images. Only 3 kernels are shown for space reasons. The blue bracketed text is the size of the images ([x-dimension, y-dimension, channel/feature dimension]). This is the exact architecture for the best performing U-Net in Section 3.}\label{unet_arch}
\end{figure*}

Even though the ANNs and CNNs described above can do skillful meteorological tasks, their architecture is best suited to do a single output (i.e., one dimensional output) like diagnosing how many lightning flashes are in a satellite image or labeling a radar image as a squall line or supercell. An advancement to neural network architectures came from \citet{Ronneberger2015}, where an architecture named \textit{"U" Network (U-Net)} was introduced. Originally designed to label and track biological cells in microscope imagery, this method lends itself to doing a valuable task called \textit{image-to-image translation}. Image-to-image translation is an example of taking some input image like infrared brightness temperature and translating it into a map of lightning data. The primary advantage of U-Nets is it will produce an image with a similar shape to its inputs. 

An example U-Net is shown in Fig. \ref{unet_arch}. The name U-Net stems from the general "U" shape the network is built in. To be clear, a U-Net is a specific type of CNN, so it contains the same makeup of convolutional layers and pooling layers of a CNN, but the U-Net differs in that it contains a series of upsampling or unpooling layers (i.e., opposite of the pooling, increasing the resolution through some interpolation technique [e.g., nearest neighbor]) instead of the ANN added to the end of the CNN that was shown in Fig. \ref{convolution_arch}.

Each step down the left-hand side in the "U" (Fig. \ref{unet_arch}) symbolizes the pooling reduction of image resolution. Then at the bottom of the "U", instead of doing additional pooling or flattening the data to be fed into an ANN (like a CNN), the data are upsampled (i.e., re-sampled to include more pixels using an interpolation method like nearest neighbor) and convolved. Then the new higher resolution images are concatenated (i.e., combined) with the same shaped images from the left-hand side of the architecture (see matrix sizes in Fig. 6) and then passed through a convolution but this time the number of filters is halved as opposed to the number of filters doubling on the left side of the "U". The concatenations from one side of the architecture to the other are called \textit{skip-connections}. The process of upsampling and concatenating is repeated until you reach the original input image shape. The left side of the 'U' is often called the \textit{encoding branch} while the right side is often called the \textit{decoding branch}. 

The intuition behind U-Nets is similar to CNNs where convolutions are used to extract spatial information. The added complexity of a U-Net beyond a CNN allows a machine learning method to produce a whole map (i.e., matrix) of predictions, instead of a single pixel or value (i.e., scalar). This is extremely useful for meteorological datasets since often in meteorology users are interested in spatial distributions of variables (e.g., dryline location). Given that forecasters have deemed timeliness an important property of machine learning meteorological tools \citep{Harrison2022}, the production of a map from U-Nets is helpful because a CNN trained to do the same task as the U-Net will require N more iterations (e.g., more time) to produce the same map, where N is the number of pixels in the map. 

Examples of U-Nets in meteorology include automatic detection of cyclones in satellite imagery \citep{Kumler-Bonfanti2020}, translating geostationary satellite data into radar data \citep{Hilburn2021}, short-term forecasts of lightning \citep{Zhou2020,Cintineo2022}, and convection \citep{Lagerquist2021a}, labeling bow echoes within model data \citep{Mounier2022} and downscaling (i.e., statistically increasing the resolution) of coarse numerical weather prediction data \citep{Sha2020a,Sha2020b}. 

\subsubsection{Summary of all machine learning methods}

\begin{figure*}[t]
 \centering
 \noindent\includegraphics[width=5.25in]{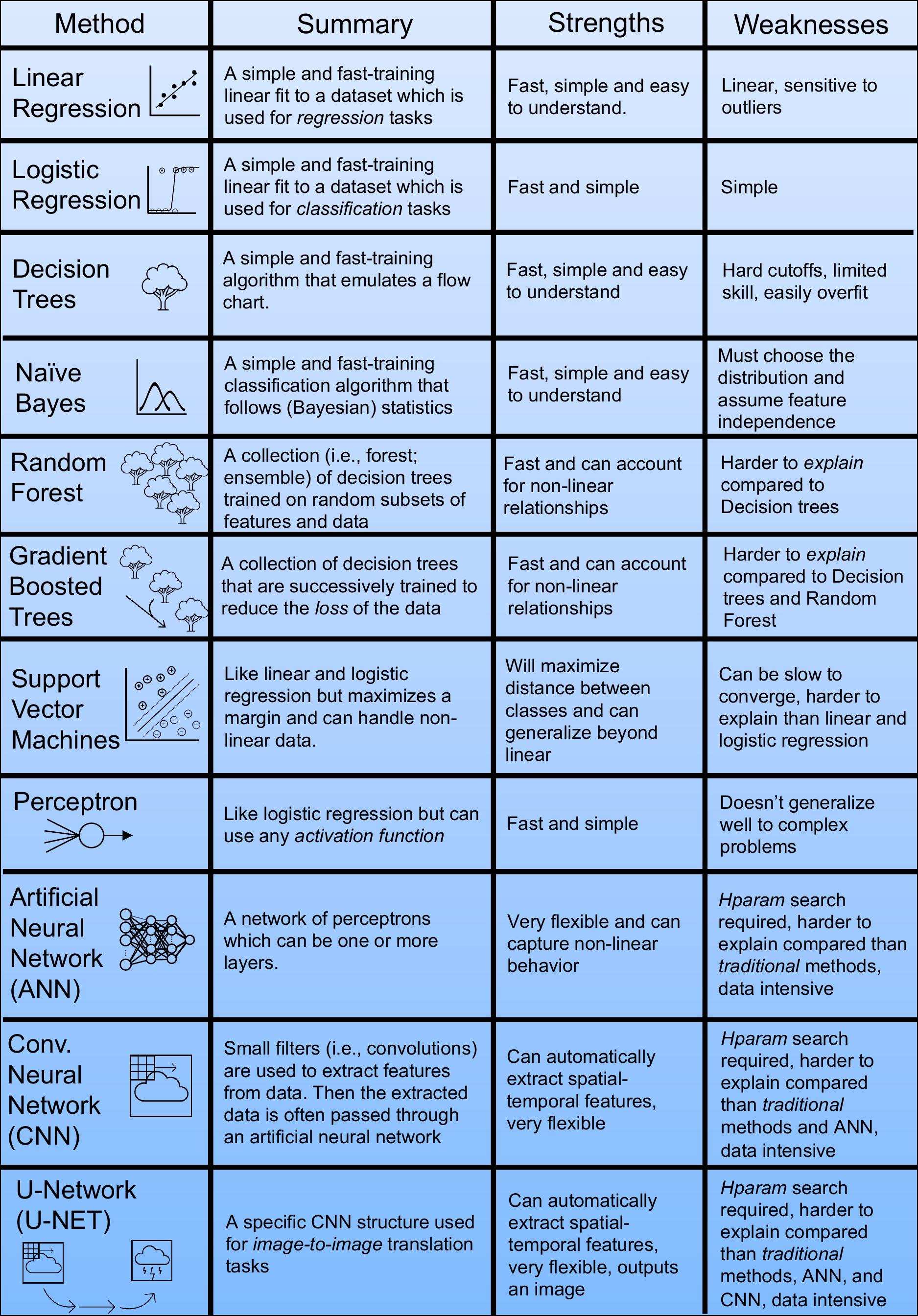}\\
 \caption{Summary graphic describing all methods discussed in both parts of this tutorial paper. The complexity of the methods increase further down the table. Hparam is an abbreviation for hyperparameter tuning}\label{summary}
\end{figure*}

By this point in the paper series (Part 1 and Part 2 combined) there have been discussions about a total of 11 machine learning methods. Thus, in order to organize and summarize these various techniques and their distinctions, Fig. \ref{summary} is provided. In the graphic there is a brief summary of each method, some strengths and some weaknesses.

\subsection{How to train neural networks \label{sec:training}}

\subsubsection{Learning the weights}

In order to determine the weights of a neural network (i.e., $\theta$) the method is similar to the traditional machine learning methods mentioned in Part 1. More specifically, the training data are used to learn the weights of the machine learning model such that the loss (i.e., error or cost) is minimized through the use of derivatives (i.e., gradients). While this simplified intuition works well for the traditional machine learning models described in Part 1, neural networks require a bit more description so that readers can navigate common vocabulary and methods that would be found in a paper describing a meteorological neural network. 

Before \citet{Rumelhart1986}, a roadblock with neural networks was the efficient and timely training of a neural network with more than a few neurons (i.e., computation took too long). As a solution, \citet{Rumelhart1986} introduced an algorithm named \textit{backpropagation} to solve for the weights of an ANN. Backpropagation works by sequentially feeding each training data example through the network, calculating the error and then calculating the change in error with respect to each of the weights, also known as the \textit{gradient} (i.e., derivative of $loss$ with respect to $\theta$). Readers can think about this gradient as the quantitative amount to change the weights in the network such that error on that example is reduced. After the gradient is calculated the algorithm adjusts the weights of the network by following a \textit{gradient descent} step: 

\begin{equation}
    \theta_{i+1} = \theta_{i} + \eta \frac{d(loss)}{d\theta}, \label{e4}
\end{equation}
where $\theta_{i+1}$ are the updated weights,  $\theta_{i}$ is the previous weights, $\eta$ is the \textit{learning rate} and $\frac{d(loss)}{d\theta}$ is the gradient of the error. The learning rate is a scalar value (e.g., $10^{-3}$) which tells the algorithm how large of a step to take. 

\begin{figure*}[t]
 \centering
 \noindent\includegraphics[width=6in]{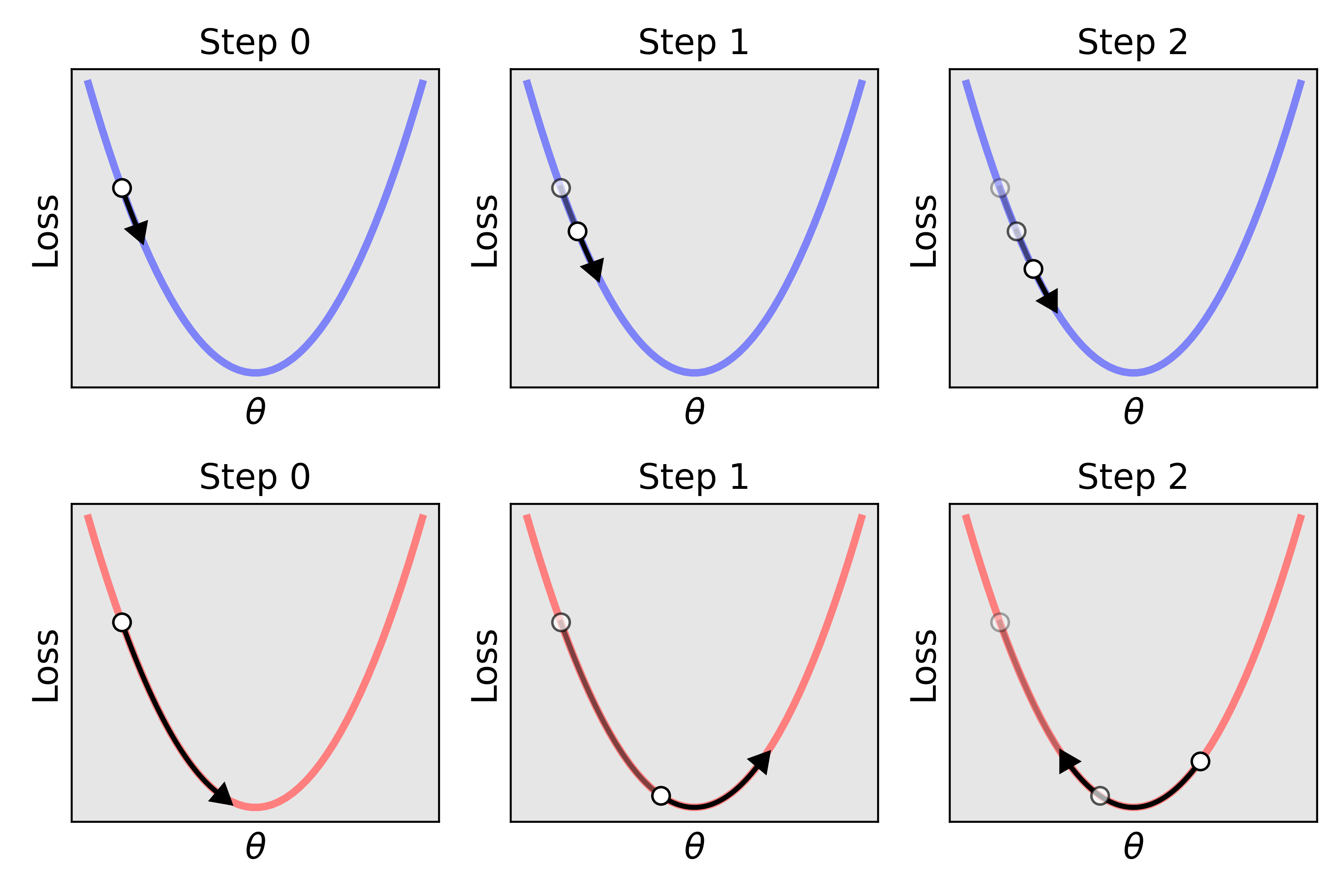}\\
 \caption{Schematic depicting gradient descent. Top row (blue colored lines) is using a smaller learning rate ($\eta$) than the bottom row (red colored lines). Arrows depict where loss will be after a gradient step. $\theta$ is the symbol representing the neural network current parameters (i.e., weights and biases).}\label{gradient}
\end{figure*}

To help illustrate this algorithm, consider the top row example visualized in Fig. \ref{gradient}. Envision the loss function on some dataset is a parabola and the neural network at the start (before any training) has a loss marked by the circle marker in the subplot labeled Step 0. After seeing a data example and calculating the gradient, the algorithm takes a step (the size of the step is determined by the learning rate) which results in the updated and lower loss in the Step 1 subplot. This is repeated, resulting in the subplot labeled Step 2. Eventually with enough steps the loss should be at a minimum (bottom of the parabola). The top row of Fig. \ref{gradient} depicts an appropriate learning rate for this example. In scenarios where the learning rate is too large, like the bottom row in Fig. \ref{gradient}, this algorithm could end up overshooting the minimum and never converging to the smallest loss. Conversely if the learning rate is too small (not shown), the algorithm will take too long to converge on the minimum loss. For these reasons, the learning rate is a hyperparameter\footnote{recall hyperparameters are parameters that are varied during training to find the best set. See Part 1 for more discussion on hyperparameters.} that is varied when training neural networks. 

Unfortunately in practice, calculating the gradient on every single training example can be too costly since the typical number of training examples is on the order of thousands to millions and the number of weights in a neural network can be similarly large. Thus, a trick around this is to use something called \textit{stochastic gradient descent}. The idea is instead of calculating the gradient and updating the weights after seeing each example, a random \textit{batch} (i.e., small collection, sub-set) of examples is used to estimate the gradient which is then used to inform the algorithm how to change the weights. The size of the batch, like the learning rate, is another hyperparameter of neural networks. The new procedure is then, select one random batch, send the batch of data through the network, calculate the loss, calculate the mean gradient of the batch and update weights (i.e., take step according to Eq. \ref{e4}). This sequence is repeated until all training data have been sent through the neural network. After the entire training dataset has been passed through the network, the network has been trained for one \textit{epoch}. Neural networks are often trained for many epochs (e.g., 50, 100, 1000 etc.) usually until the loss doesn't change much (i.e., changes less than $10^{-6}$)\footnote{this can be user defined, but $10^{-6}$ is a common choice} or when over-fitting is detected.

In practice, stochastic gradient descent is just one method of \textit{optimizing} a neural network. Other \textit{optimizers} can be used to train neural networks, but for the sake of this tutorial, they all generally follow the same steps as stochastic gradient descent. The names of other popular optimizers a meteorologist might encounter are: the adaptive moment estimation \citep[Adam;][]{adam} and root mean square propagation \citep[RMSprop;][]{Tieleman2012}. 

\subsubsection{Loss functions}
Just like the traditional machine learning methods, neural networks can be used for both categories of supervised machine learning: classification and regression. The primary differences between a neural network for classification and a neural network for regression is which loss function is optimized and what output activation is chosen (i.e., activation of the last node or layer). For classification, typical loss functions include \textit{binary cross-entropy} and \textit{categorical cross-entropy} accompanied with a sigmoid (see Fig. 3 and Eq. 6 in Part 1) or softmax (a variant of sigmoid) output activation function for binary and multi-class classification tasks respectively. Meanwhile for regression, common loss functions include mean absolute error and mean squared error accompanied with a linear output activation. More sophisticated loss functions can be used, like the fractions skill score \citep{Roberts2008}, and are an active area of research within machine learning for meteorology \citep{Ebert-Uphoff2021,Lagerquist2022}.

\subsubsection{Regularization and Over-fitting}
Neural networks can often contain hundreds, thousands or even millions of trainable parameters. While this enables neural networks to be very flexible, it can also enable the network to over-fit to the training data very easily. Thus, there are some specialized methods that can help prevent overfitting (i.e., regularize) the neural network. A popular method of regularization is called \textit{dropout} \citep{JSrivastava2014}. Dropout is where neurons within a layer of the network are randomly turned off (set to 0) in the training process. The neurons that are turned off are changed after each batch, but the percentage of neurons turned off in the layer is constant over the training time and is a hyperparameter choice (e.g., 10$\%$ of neurons). Then when the model is used in \textit{inference mode} (i.e., prediction time) the dropout layers are not used, meaning all the neurons are used. The intuition behind dropout is that if random neurons are turned off during training, the network is forced to learn redundant pathways and cannot simply memorize a single pathway through the network for each example it sees.  

A second regularization method commonly used is called \textit{data augmentation}. Data augmentation are synthetic alterations made to the training data. These alterations include things like, random rotations, random flips (up-down or left-right or both) and adding random noise. The reason this is done is because adding these slight alterations provides the neural network with slightly different examples to learn from, which in turn makes your neural network model more resistant to overfitting and more robust to things like measurement errors. Data augmentation is also a way to increase your training sample size without having to actually add more data. 

A third method of regularization is called \textit{batch normalization} \citep{Ioffe2015}. Batch normalization, as the name suggests, normalizes the values of a batch of data within the neural network. The reason for this stems from the use of batches themselves, which are needed for timely training of neural networks. Because the training process randomly selects a batch of data to estimate the gradient from, that batch of data is not guaranteed to have properties that are well suited for stable training, like being normally distributed. Thus, to assure that training goes as smoothly as possible, batch normalization layers can be inserted after any layer in a neural network. 
 
\subsubsection{Hardware} 
A meteorologist will likely encounter discussions of what hardware (i.e., computer details) are being used to do the neural network training. This discussion comes from the issue that training a neural network can be computationally very slow on a normal computer (i.e., central processing units [CPU]). As a way to speed things up, the open-source neural network software packages, named Tensorflow \citep{tensorflow2015-whitepaper} and PyTorch \citep{pytorch}, have built their software to allow users to utilize a computer chip called a Graphical Processing Unit (GPU). The GPU enables the calculation of the convolution of an image and the gradients to be much faster, which ultimately accelerate training. While there are many different types of GPUs and CPUs and many different neural network tasks, in general a GPU can often reduce training time by a factor of two to ten. The Google Colab notebooks (see data availability section) that accompany this manuscript leverage the freely available GPUs provided by Google in the cloud. 


\section{Neural network application and discussion \label{sec:ml_app_dis}}


\subsection{Problem Statements \label{sec:problem}}
Here we restate the machine learning problem statements explored in this paper. We again apply the Storm EVent ImagRy \citep[SEVIR;][]{Veillett2020} dataset to two main tasks: (1) Does this image contain a thunderstorm?  and (2) How many lightning flashes are in this image? For more information about the SEVIR data see Part 1. We assume the GOES Lightning Mapper (GLM) observations are unavailable and we need to use the other measurements (e.g., infrared brightness temperature) as features to estimate if there are lightning flashes (i.e., classification), and how many of them are there (i.e., regression). Both tasks (1) and (2) are centered on using machine learning models having a singular (i.e., one dimensional) output. As we mentioned in Section 2.a.4, U-Nets offer more than a single output (i.e., two dimensional), as they re-create an entire image as an output. Thus, the problem statements for the U-Net application are then: (1b) Label the pixels in this image where there are lightning flashes; and (2b) For each pixel, diagnose the number of flashes in that pixel. 

\begin{figure}[t]
 \centering
 \noindent\includegraphics[width=2in]{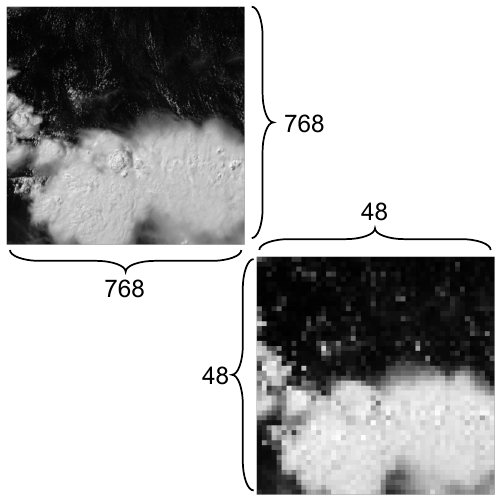}\\
 \caption{Example of the full resolution visible image (top-left) and its reduced resolution in sub-SEVIR (bottom right). The numbers correspond to the number of pixels along each dimension.}\label{resize}
\end{figure}

\subsection{Data \label{sec:data}}

Before jumping into the results of the trained neural networks, we want to emphasize an intersection between neural networks and the traditional methods discussed in Part 1. The discussion from Part 1 Section 3b regarding data curation applies to neural networks as well. Specifically, a dataset used to develop a neural network must also be split into independent subsets for training, validation, and testing the model. Thus, to follow Part 1 we use the same datasets with a slight alteration. While the original SEVIR dataset is primed for successful machine learning, the size of the raw dataset (approximately one terabyte in storage size) is cumbersome for the intents of a tutorial and could not be used on most personal computers. Thus, as an effort to make this dataset more accessible as a tutorial, we have reduced its size. To do this, we first reduced all images to have the same resolution of the gridded lightning data (48 pixels by 48 pixels;  approximately 8 km by 8km pixels). Figure \ref{resize} shows an example of the full resolution visible image and its corresponding low resolution version. After the images were re-sampled, one random continuous hour (12 images) of the four total hours (48 images) for each storm event is kept. 
Since we are keeping the same number of storm events, we keep the training, validation and testing data splits the same as Part 1, which were 01 Jan 2017 - 01 Jun 2019 for training and split every other week in the rest of 2019 into the validation and test sets. Doing both of these re-samplings of SEVIR results in a more manageable dataset (approximately two gigabytes in storage size), while also preserving 60,000 training samples and about 12,000 validation and test samples. We name this subset of SEVIR: \textit{sub-SEVIR}, and the location of the dataset can be found in the data availability section.

Owing to reduced resolution, the sub-SEVIR dataset contains different information.  Given our goal of comparing the neural network models of this paper to the machine learning methods of Part 1, we must re-extract the same features from sub-SEVIR. Specifically, we extract the following percentiles: 0,1,10,25,50,75,90,99,100. These percentiles are then used as input features for re-training the traditional machine learning methods and to serve as a baseline comparison with trained neural networks. 

\subsection{Training the networks \label{sec:train}}
After reading the section on how to train neural networks (Section 2b), the reader might notice that there are numerous hyperparameters for neural networks. In Part 1, the traditional machine learning models shown were trained with the default hyperparameter choices as defined by the scikit-learn Python package \citep[scikit-learn;][]{scikit-learn}. The idea of default hyperparameter choices does not necessarily exist with neural networks. Thus, it is good practice for those training neural networks to run some sort of hyperparameter search (i.e., vary a bunch of the parameters) because users are not guaranteed to get good performance with some starting parameter choices. For example, recall the discussion about choosing a learning rate in section 2.b.1. 

All the trained neural network models shown here are the result of a hyperparameter search. We conducted 100 random hyperparameter configurations for each neural network trained and systematically varied things like the number of layers, the number of neurons, the loss etc. In the end we chose one of the 100 models to show in the following section. These were chosen based on their performance in the validation set. For readers interested in the exact hyperparameter choices we varied to find out best performing models, see Fig. A1-A3 in the appendix.

\subsection{Classification}
\begin{figure*}[t]
 \centering
 \noindent\includegraphics[width=5.5in]{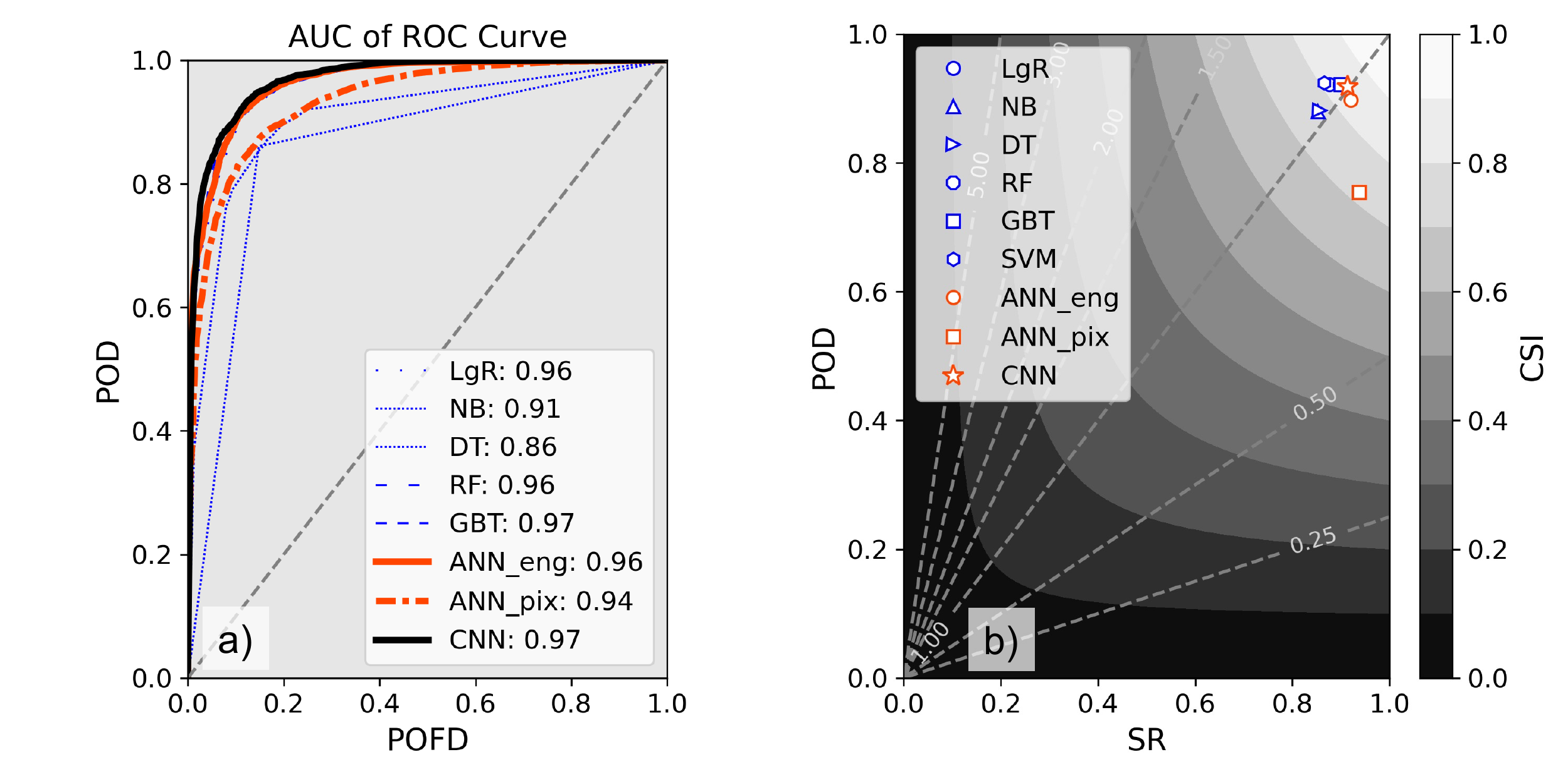}\\
 \caption{Classification Metrics (a) ROC curve diagram. All thin blue lines are the traditional machine learning methods from Part 1 (LgR: logistic regression; NB: naïve Bayes; DT: decision tree; RF: random forest; GBT: gradient boosted decision tree;). The thick lines are the neural networks trained ($\mathrm{ANN_{eng}}$; ANN trained with engineered features; $\mathrm{ANN_{pix}}$ ANN trained using pixels as features). Values in legend are the AUC values. (b) The performance diagram for each machine learning model \citep{Roebber2009}.}\label{class_performance}
\end{figure*}

The first machine learning task we consider is (1) to classify a SEVIR image if it has at least one lightning flash within it. To serve as a comparison, all of the traditional machine learning methods from Part 1 were re-trained on the sub-SEVIR dataset. Their performance on the validation dataset is shown as the thin blue lines and blue markers in Fig. \ref{class_performance}. 

For (1), we also trained three neural networks. The first is an ANN trained with the same input features (i.e., the table of percentiles extracted from each image) as the traditional machine learning models (red solid line and red circle Fig. \ref{class_performance}). This ANN trained on the engineered features (i.e., the percentiles of the image) effectively reproduces the performance of the best traditional machine learning methods (e.g., gradient boosted trees). Since there is a relatively similar performance of random forest, gradient boosted trees and ANN, it would be better to use the tree-based methods over the ANN operationally for this task. This is suggested because tree-based methods are less complex and thus more \textit{interpretable} (c.f., Figure 1 in Flora et al. 2022). Using a less complex and more interpretable model also provides a better opportunity to meet the \textit{consistency} point made by \citet{Murphy1993}. 

The second neural network trained is another ANN but this time it was trained using each pixel as a feature (e.g., Fig. \ref{pixel_input}). The reason a second ANN is trained, is to see if the ANN could learn important features on its own, without a domain scientist (i.e., meteorologist) extracting pertinent information (i.e., the percentiles from the satellite images). While the pixel trained ANN has generally good skill (AUC $>$ 0.9, CSI $>$ 0.7 red;  dash-dot line Fig. \ref{class_performance}a; red square Fig. \ref{class_performance}), the result is worse than all other methods discussed so far (Fig. \ref{class_performance}b).

The last neural network trained for task (1) is a CNN. To be explicit, recall that the CNN uses the raw images as inputs and convolves them to extract features. The result of training a CNN on the sub-SEVIR data provides one of the best performing machine learning methods (black line Fig. \ref{class_performance}a; red star Fig. \ref{class_performance}), matching the skill of the gradient boosting trees and the ANN trained on the engineered features. Note that the CNN only marginally outperforms the other methods on the performance diagram and is likely not a significant difference. 

It might be surprising to see that the ANNs do not substantially outperform the tree based methods on this task despite the added complexity of neural networks and their training. This is a common pitfall for machine learning users. In fact, there is growing evidence that the tree based methods can often outperform neural networks and deep learning on \textit{tabular data} \citep[i.e., data contained in a spreadsheet;][]{Shwartz-Ziv2022}. A distinction is made between tabular and non-tabular datasets here because spatial details can contain substantial information for the machine learning task and isn't always easily quantified into a tabular dataset. For example, consider assessing a storm's tornadic potential. While using composite radar reflectivity as a feature could be useful (e.g., strong radar value means a strong storm), there is likely more information contained in the shape of the radar echo (e.g., is there a hook echo?). Thus, given the amount of additional effort required to explore the hyperparameters in neural networks, our suggestion is that if you have a tabular dataset, start with random forest and gradient boosted trees for your machine learning model. Often times this will result in a useful machine learning model without the headache of doing a large hyperparameter search or needing specialized computers (i.e., GPUs). Otherwise, if you have a spatial dataset (e.g., radar images) and you are unsure of what features to extract, then the extra effort of CNNs could be beneficial. 

\begin{figure*}[t]
 \centering
 \noindent\includegraphics[width=5.5in]{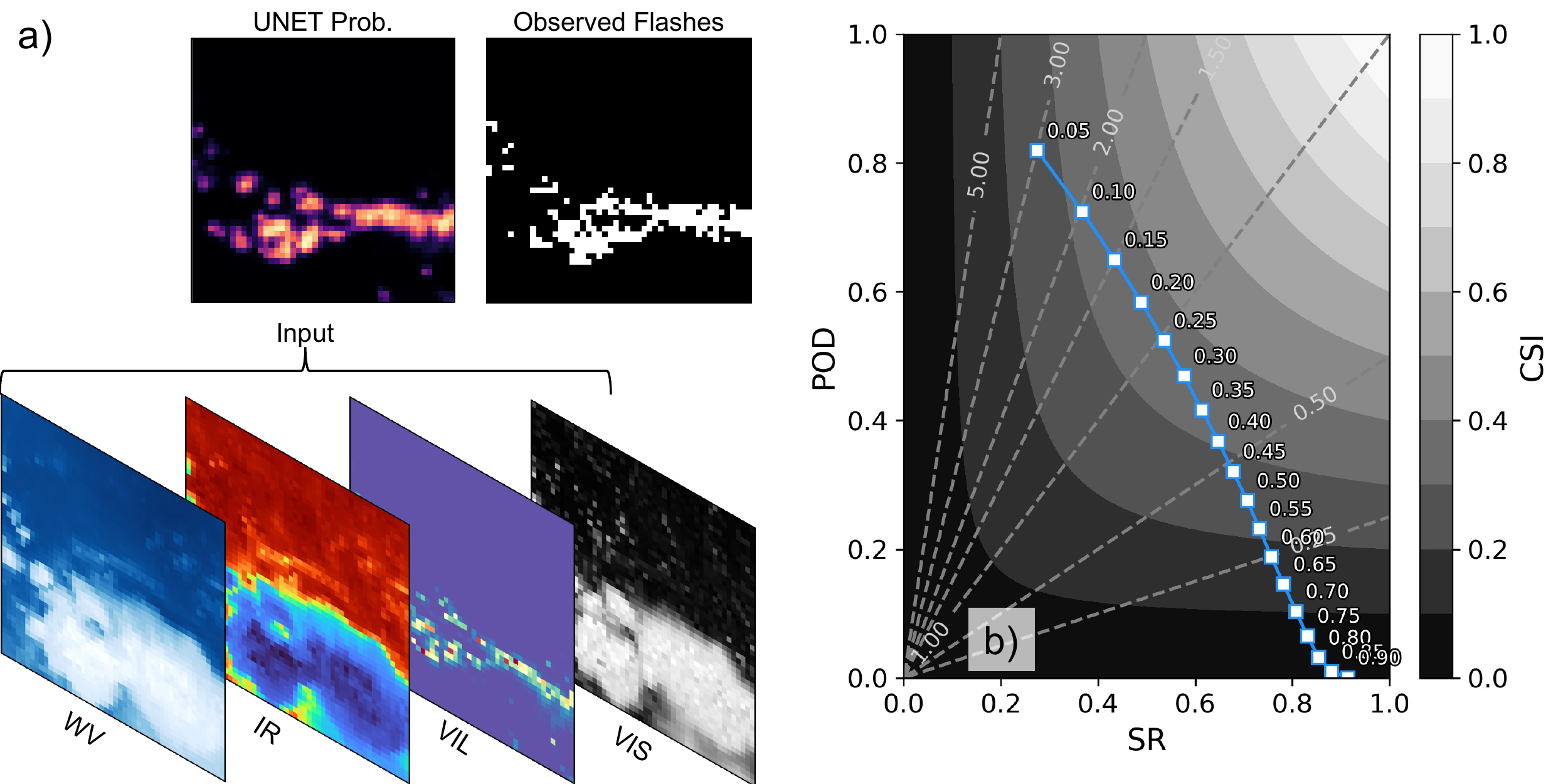}\\
 \caption{Trained U-Net results for classification. (a) Example input and output of the U-Net. (b) Performance diagram for the U-Net evaluated on every pixel. The numbers next to the markers show the probability threshold to classify a pixel as containing lightning or not.}\label{class_unet}
\end{figure*}

Moving beyond the single output models, a U-Net for classification is also trained and evaluated. Recall that a U-Net (e.g., Fig. \ref{unet_arch}) outputs a map with the same shape as the input images. In other words, a U-Net tasked with problem statement 1b (in Section 3.a) produces an output map where each pixel is assigned a probability of it containing lightning. An example output is shown in Fig. \ref{class_unet}a. A U-Net can be evaluated similarly to the previous models and a performance diagram for the trained U-Net is shown in Fig. \ref{class_unet}b. Note on Fig. \ref{class_unet}b that instead of a single marker the figure shows a line with many markers. This is because the threshold for deciding if a pixel is labeled as no lightning or lightning is varied from zero to one at 0.05 increments. This is done because a model could potentially get better results if a probability threshold other than 0.5 is used (which was shown in Fig. \ref{class_performance}), which is the case for this U-Net. Note that this could also be done with the all of the other machine learning methods (except support vector machines) shown in Fig. \ref{class_performance}b, but the threshold of 0.5 generally works well for those models. 

Comparing Fig. \ref{class_performance}b and Fig. \ref{class_unet}b, initially it seems like the U-Net is performing worse than the ANNs and CNN because the line on Fig. \ref{class_unet}b is well below the location of all other models in Fig. \ref{class_performance}b. That being said, it is unfair to compare the two sets of performance statistics because the U-Net is being evaluated on every single pixel rather than on the image as a whole. Given the added complexity in Problem Statement 1b, the U-Net performance is encouraging with CSI values of 0.36 when using a probability threshold of 0.25. This offset from the probability threshold of 0.5 happens frequently in meteorology and can be mostly attributed to rare phenomena and the training dataset being imbalanced. 

It might not seem like the lightning flashes are rare but if you consider the total number of pixels that contain lightning they make up less than one percent of the total amount of pixels. Thus, given the number of no lightning pixels far out weigh the lightning pixels, the U-Net will learn this natural distribution and skew its output to account for the more likely outcome. The result is that on the performance diagram, a lower probability threshold can perform better than using the default 0.5. While altering the probability threshold for pretrained models can improve performance, other mitigation techniques can be taken and are focused on adjusting the ratio of non-zero pixels to zero pixels in the training dataset. 

One way to adjust the ratio of pixels is by subsampling the 48 by 48 pixel images into smaller \textit{patches} (e.g., 24 by 24) and only train on patches that have a larger proportion of non-zero pixels. This tends to work well, but is more resource intensive because patching the data requires the user to then \textit{stitch} the patches back together while using the model output. Another way would be to adjust or change the loss function to weight the classes differently. By default most loss functions weight all classes equally. There are ways to adjust the loss function and tell your machine learning model that the rare classes are more important that the training data suggests. For examples of custom loss functions see \citet{Ebert-Uphoff2021}. Alternatively, one could do both subsampling and a differently weighted loss function. These alterations can be considered part of the hyperparameter tuning of the U-Net training. A meteorological example of exploring various U-Net training procedures can be found in \citet{Mounier2022}, where a U-net is used to identify bow echoes. Note that weighted loss functions and resampling the training data is not exclusive to U-Nets. These methods can be explored for all neural networks.

\subsection{Regression}

\begin{figure*}[t]
 \centering
 \noindent\includegraphics[width=5.5in]{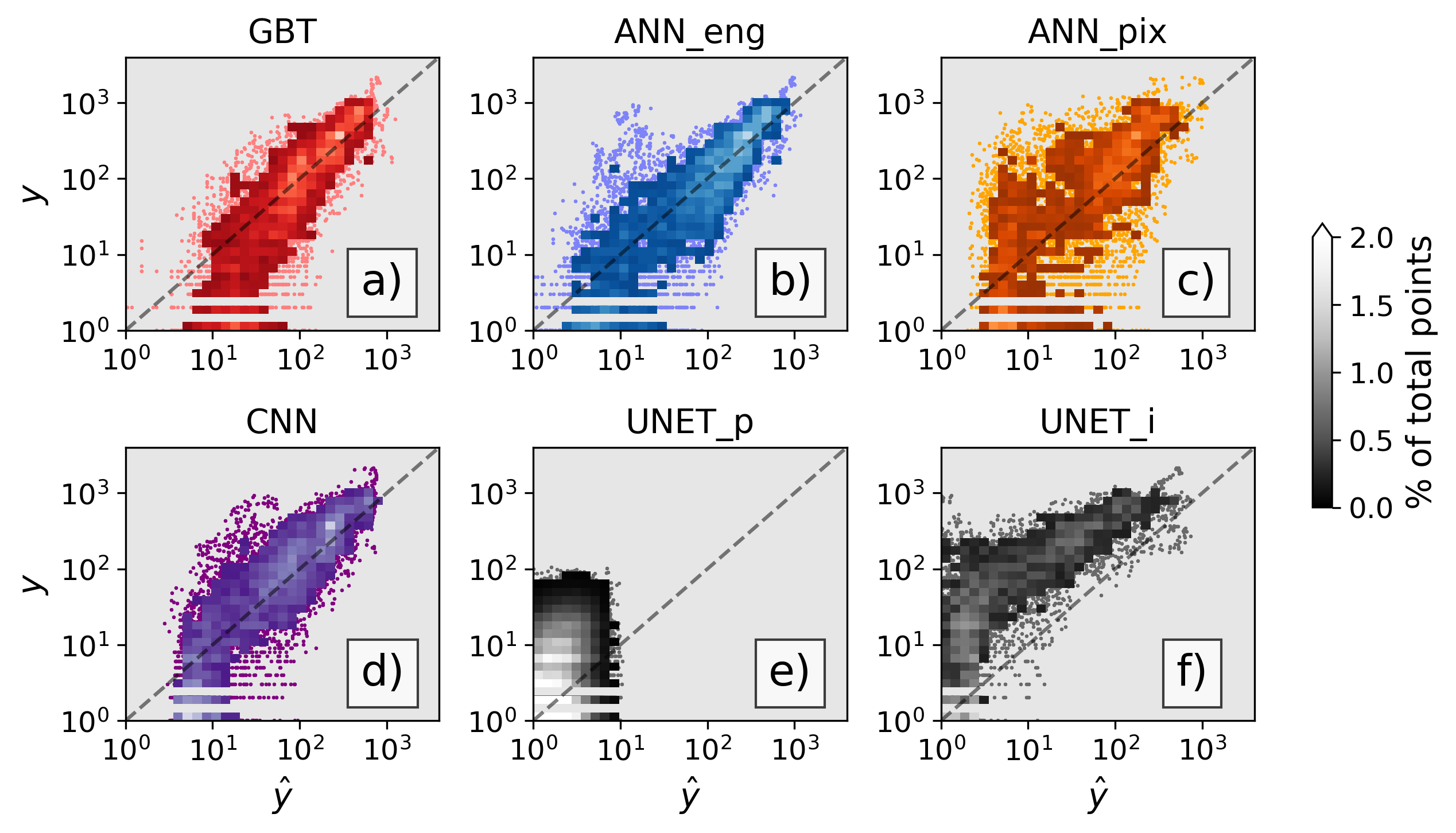}\\
 \caption{One to one diagrams with all regression methods trained on sub-SEVIR. The x-axis is the machine learning model prediction ($\hat{y}$) and the y-axis is the truth ($y$): (a) Gradient boosted trees (b) Artificial neural network using the tabular data (c) artificial neural network using the pixel data (c) Convolutional neural network (d) U-Net evaluated on every single pixel in the images (e) The U-Net evaluation but on the sum of all lightning flashes in an image.}\label{reg_onetoone}
\end{figure*}

\begin{figure}[t]
 \centering
 \noindent\includegraphics[width=2.5in]{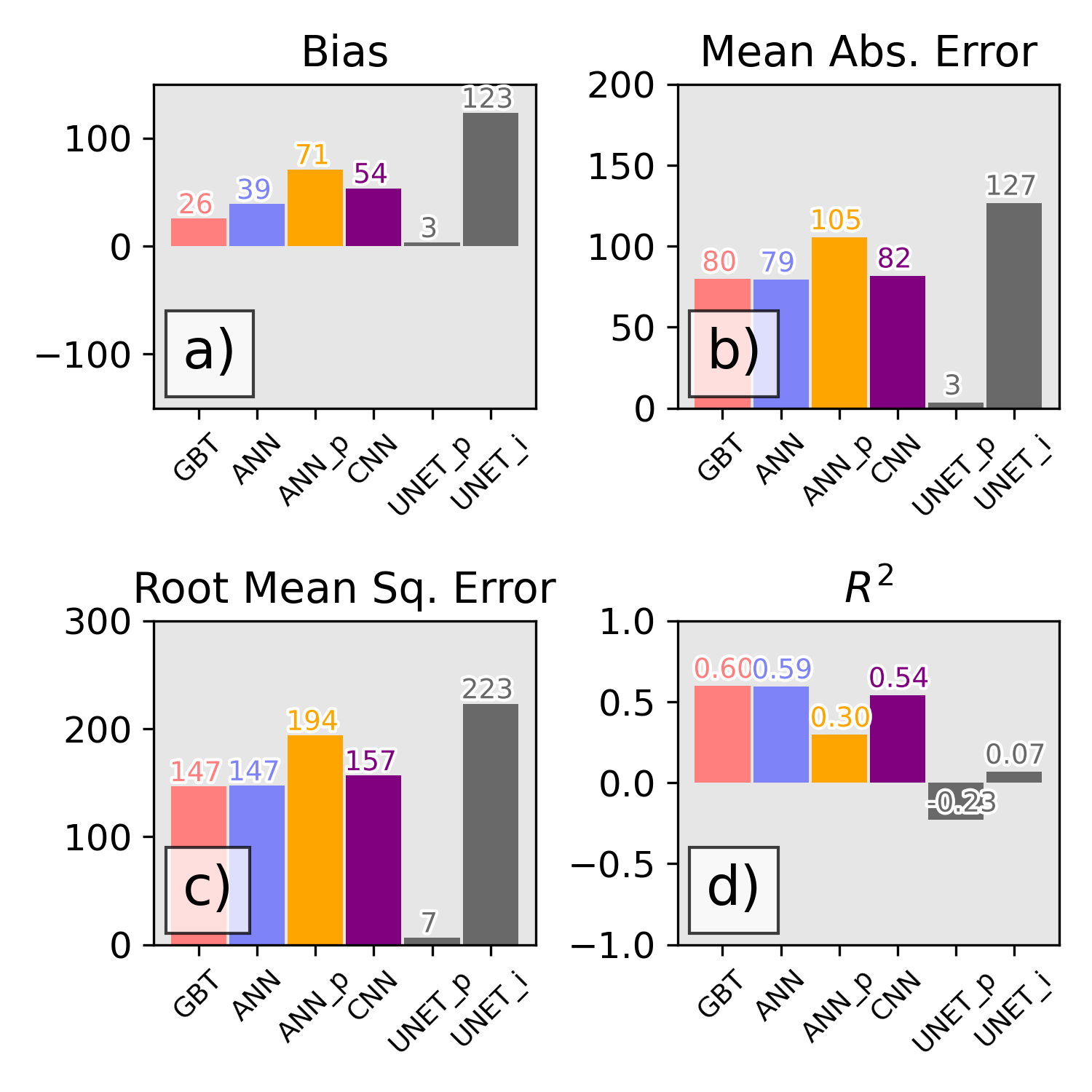}\\
 \caption{Metric bar charts with all regression methods trained on sub-SEVIR}\label{reg_metrics}
\end{figure}

Considering task (2), the goal of the machine learning is to now determine the number of flashes that are occurring in a SEVIR image. Like the previous section, the goal is to compare the neural network methods presented in this paper to the traditional machine learning methods of Part 1. To make the comparison more concise, we only show the best performing regression model trained on the sub-SEVIR dataset, which was the gradient boosted trees. Recall that for this regression task, only data examples that had more than one flash in them were used as the training data. The performance of the gradient boosted tree on the validation dataset is shown in Fig. \ref{reg_onetoone}a and the red bar in Fig. \ref{reg_metrics}. 

For (2), a similar suite of neural networks as the classification task are trained and their performance is characterized in the same way as regression in Part 1. The first neural network trained is the ANN using the engineered features as inputs. Akin to the results of the classification task, this ANN achieves similar performance to the gradient boosted trees. The ANN has a high density of points that follow the diagonal in Fig. \ref{reg_onetoone}b and has a mean absolute error, root mean square error and $R^{2}$ values very close to the gradient boosted tree (blue bar Fig. \ref{reg_metrics}). That being said, the bias of the ANN is larger than the bias of the gradient boosted tree (Fig. \ref{reg_metrics}). 

A second neural network trained is an ANN using the pixels as features. It is clear that this model has issues. The points on the one-to-one plot are more spread out and not highly concentrated along the diagonal (oranges Fig. \ref{reg_onetoone}c). All metrics are worse compared to the ANN and gradient boosted tree trained on the engineered features. This result is very similar to the classification model, where the model has some skill but performance is considerably worse when the ANN has to learn what features are important based on the pixels as input. 

A third network trained is a CNN. The CNN achieves similar performance to the ANN trained on the engineered features and the gradient boosted tree. The points are more densely aligned along the diagonal in Fig. \ref{reg_onetoone}d (purples) and the quantitative metrics (purple bar Fig. \ref{reg_metrics}) are effectively the same as the ANN, but it does have worse bias (Fig. \ref{reg_metrics}a). Thus, like the classification task the CNN was able to extract relevant features to make a skillful designation of the number of lightning flashes in the image.

A regression U-Net is also trained and evaluated. Instead of determining the probability of lightning in each pixel, the designation of the regression U-Net is the number of flashes in each pixel. A similar problem occurs with the regression U-Net as with the classification U-Net when trying to compare the U-Nets to the other neural networks. Consider the pixel-wise evaluation of the regression U-Net (Fig. \ref{reg_onetoone}e and Fig. \ref{reg_metrics}). The U-Net has a clear underestimation of the number of flashes compared to the observed flashes and yet the mean absolute value, bias and root mean square error are close to zero. This evaluation might initially seem contradictory, but the pixel-wise distribution of flashes is two orders of magnitude smaller than the image wise number of flashes (e.g., mean pixel number of flashes is 3 while mean image number of flashes is 150). Since the magnitude is smaller, the metrics are correspondingly smaller. Thus, the comparison of this U-Net to the other neural networks is not necessarily fair.

As an alternative evaluation, the sum of U-Net predicted flashes across all pixels in an image can be calculated. The sum of all flashes in the images results in approximately an order of magnitude offset in the designation (Fig. \ref{reg_onetoone}f). Like the underestimation of the U-Net in the classification example, the regression underestimation probably occurs because skew of the distribution of pixels with lightning and pixels without lightning. The regression example is further compounded by the strong left skew (i.e., toward zero flashes) in the distribution of pixels with lightning flashes. The previously discussed mitigation techniques for the classification U-Net can also be applied to regression (e.g., changing the loss, patching etc.). 

\subsection{Explainable Artificial Intelligence (XAI)}

As mentioned in the motivation of Part 1, machine learning methods are often seen as \textit{black boxes} where the user cannot see what the machine learning is using to make its decisions and predictions. To combat the opaqueness of machine learning methods we present two methods, permutation importance and accumulated local effects, which can be applied to the traditional machine learning methods that made the black box more transparent. Here we show something similar but applied to neural networks. In machine learning, the methods used to \textit{explain} a machine learning output are commonly refereed to as eXplainable Artificial Intelligence (XAI). The XAI field is a place of active research development and readers can see \citet{Flora2022a,Flora2022b} for additional discussion of XAI techniques for the traditional machine learning methods and \citet{Mamalakis2022,Mamalakis2022b} for XAI techniques for neural networks. Know that the following discussion and examples only show the XAI techniques applied to CNNs but these techniques can be applied to all the neural networks discussed in Section 3.  

\subsubsection{Permutation Importance}

\begin{figure}[t]
 \centering
 \noindent\includegraphics[width=3in]{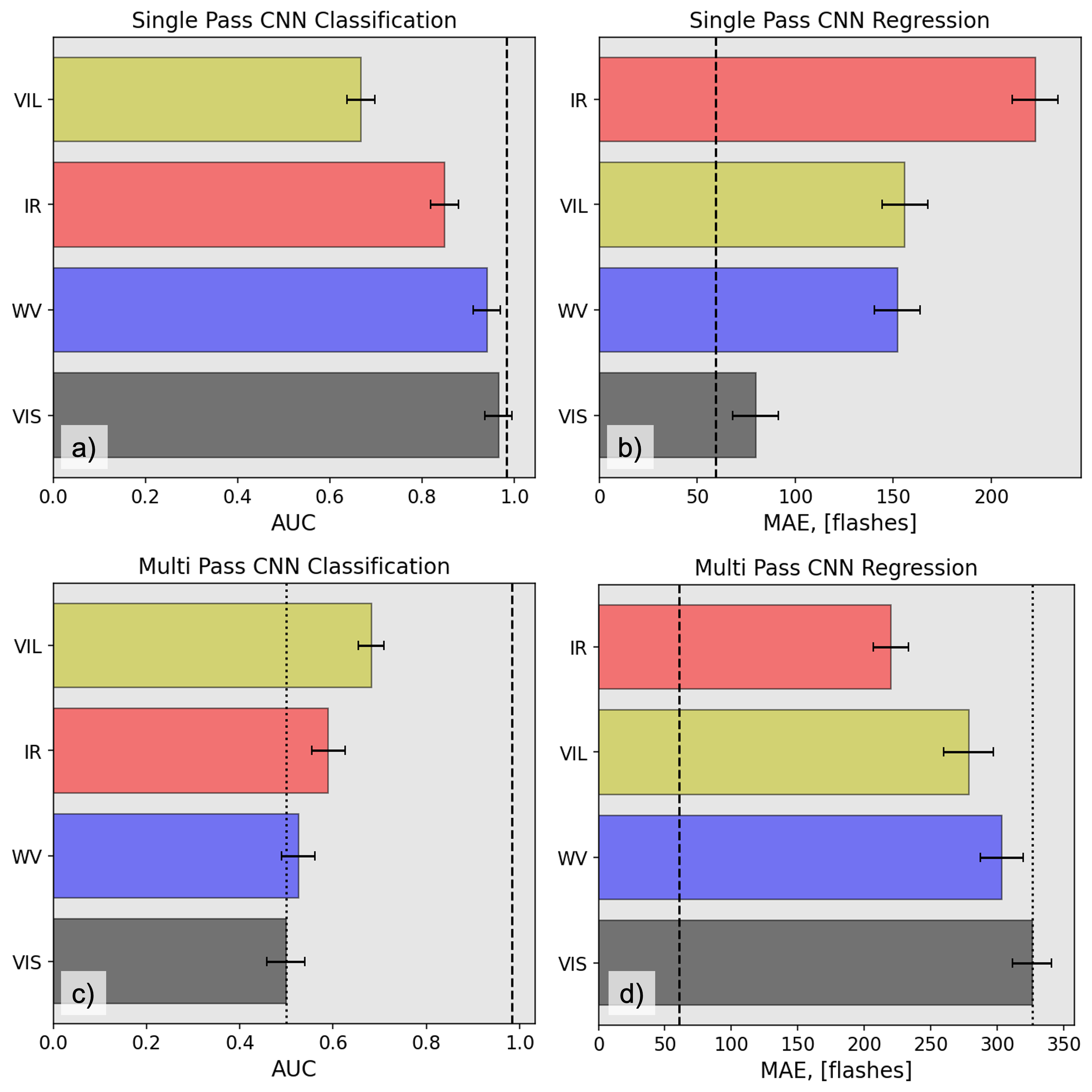}\\
 \caption{Permutation Importance results of the convolutional neural networks for both classification and regression. Yellow bar is the vertically integrated liquid, red bar is the infrared, blue is the water vapor and black is the visible. Top row (a,b) are single pass backward results and the bottom row (c,d) are multi-pass backward results. The left column is for classification (a,c) and the right column is for regression (b,d). All results are computed on 30 random samples of 250 images from the training dataset. The dashed line is the original score before shuffling any features, while the dotted line is the final score for shuffling all variables (only multi-pass)}\label{perm_imp}
\end{figure}

The first XAI method shown here is the same as Part 1, Permutation Importance \citep[][]{Breiman2001, Lak2015}. We show this method because it is a powerful method than can help users understand which inputs to the machine learning model are most important. Also, we choose to show this technique because of how flexible the method is to be used on any machine learning method. 

The general procedure is the same as discussed in Part 1 (Section 3.c.1). Input features are shuffled one by one, such that the change in the desired metric quantifies the feature's importance to the machine learning model (i.e., single pass). Since we are doing this technique on images, the difference from Part 1 is that first the pixels within an image are shuffled, then the order of images are shuffled to properly make the input features random. From there the procedure is exactly the same as in Part 1.  

Figure \ref{perm_imp} is an example of permutation importance applied to both the CNN for classification and the CNN for regression. The interpretation of this figure is the same as Part 1, but now features are grouped grouped according to the variable from which they originated. For example, the single pass result shows that the vertically integrated liquid is the most important feature for diagnosing if an image has at least one flash in it, while the infrared channel is the most important feature for determining the number of flashes in an image. For this example, the multi-pass method show the same result, but know that this is not always the case. 

These designated important features make sense meteorologically. Vertically integrated liquid can be interpreted similarly to radar reflectivity. If one where to look at an image where there is no radar reflectivity measured, it would be simple to say there is no lightning in the image. Meanwhile, since the regression task is evaluated on only examples that have at least one flash, the model is leaning more heavily on infrared. This could be because the amount of cold cloud tops in an image is plausibly related to how much lightning is in the image (e.g., more updrafts can lead to more clouds which could lead to more lightning), but further testing would need to be done to confirm or deny this explanation of the machine learning reasoning. 

\subsubsection{Deep SHAP}

\begin{figure*}[t]
 \centering
 \noindent\includegraphics[width=6in]{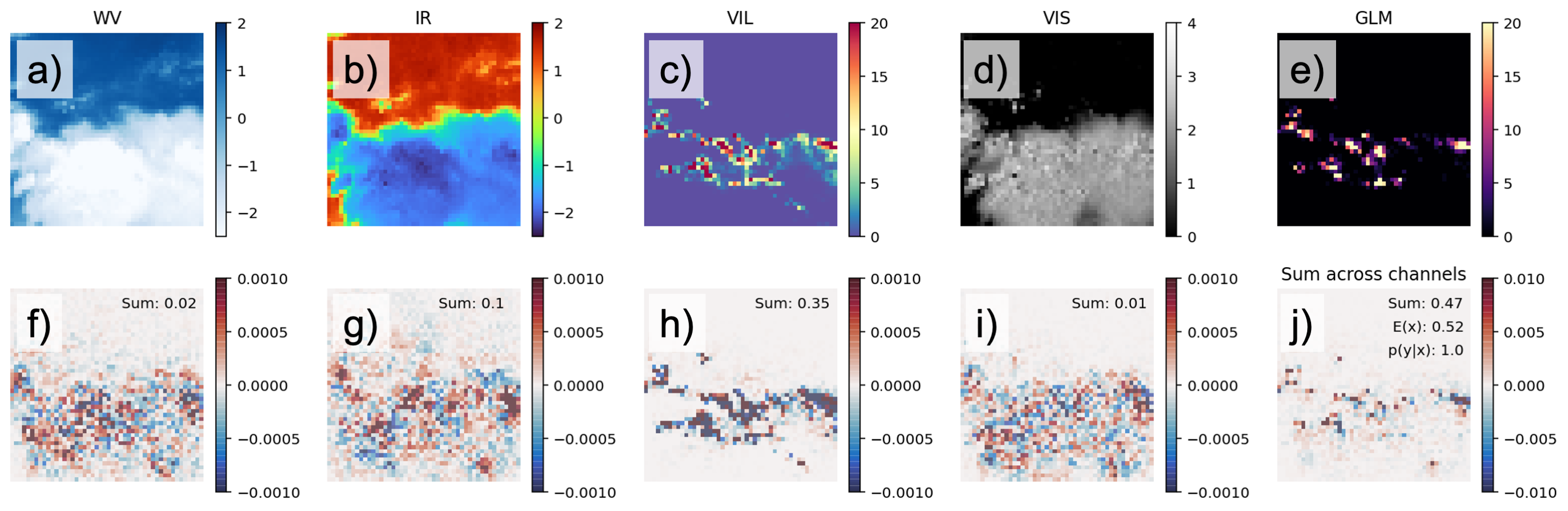}\\
 \caption{Deep SHAP estimation of SHAP values for a case in the SEVIR validation dataset (2019-08-19).(a) water vapor (b) infrared (c) vertically integrated liquid (d) visible. Note that values for panels a-d are scaled and are thus unit less.(e) GOES lightning mapper number of flashes in each pixel. (f-i) SHAP values for each respective channel of the input. The sum of all SHAP values in the image is annotated in the top right corner. (j) Sum of all SHAP values across the channels with the total sum, expected value and the ML output written in the top right corner.}\label{shap}
\end{figure*}

The second XAI method we discuss is called Deep SHAP (Lundberg and Lee, 2017), which estimates Shapley values \citep[i.e., SHAP values;][]{Shapley1953} that quantify the effect each input feature contributes to the total machine learning model output. SHAP values are calculated using a branch of mathematics called game theory, which enables the SHAP values to consider interactions between features (e.g., water vapor is correlated to infrared brightness temperature) while also allowing timely computation. While understanding how SHAP values are exactly calculated can be complicated, their interpretation is relatively straightforward and have some attractive properties. 

Consider an example of SHAP values for the classification CNN model on one of the examples (Fig. \ref{shap}). We can see in this example that there is deep convection in the bottom half of the image, characteristic of cold cloud tops (Fig. \ref{shap} ab), large vertically integrated liquid values (Fig. \ref{shap}c) and lots of observed lightning (more than 1000 flashes in this 5 min observation; Fig. \ref{shap}e). Using DeepShap, the estimated SHAP values for each feature is shown in the corresponding image below the input data (Fig. \ref{shap}f-j). The way to interpret SHAP values are that negative values (blue colors in Fig. \ref{shap}f-j) have negative attribution, or contribute negatively to the output (i.e., evidence against lightning in the image), while positive values (red colors in Fig. \ref{shap}f-j) have positive attribution (i.e., evidence for lightning in the image). 

A general interpretation of the SHAP values in Fig. \ref{shap} is that the ML model is using pixels where there are clouds for its output (i.e., SHAP colors show up where cloud is). While this might seem like an unimportant result, it is never guaranteed that the ML model will use logical decision techniques. There have been notable examples in the computer science literature where the ML identifies unexpected parts of an image to do its output, like a copyright symbol or a company logo \citep{Lapuschkin2019}. It is then encouraging that the ML is considering the clouded part of the image to diagnose if there is lightning within in. Another interpretation from the SHAP values is that the clouded region contributes both positively and negatively to determining if there is lightning in this image. This decision making process is not expected but could be a result of the ML task of determining if there is at least one flash in the image and not determining \textit{where} in the image the lightning are. Since the ML task is for the entire image, then the SHAP values should also be interpreted more holistically where the sum of the SHAP values across the clouded area can be compared against the sum outside the clouded area, where the sum is larger in the clouded region (i.e., more red than blue).  

The summation of SHAP is enabled by its additive formulation. By design the SHAP values, when added to the expected value (i.e., mean output from all images) results in the output of ML model. This additive property enables more than the discussion above of the clouded and non-clouded region, but also the relative importance of every input channel to the output of the ML. For example consider the channel wise SHAP sums in the top right corner of Fig. \ref{shap}f-i. The SHAP values for vertically integrated liquid is the largest sum with a value of 0.35, followed by infrared brightness temperature with a value of 0.1 and then water vapor and visible with values of 0.02 and 0.01 respectively. This is a similar result to the permutation importance result which provided evidence that the vertically integrated liquid is the most important input variable. Lastly if you consider some of the SHAP values in a pixel-wise sense (Fig. \ref{shap}j), the SHAP values mainly outline the edges of the vertically integrated liquid input channel. 

\begin{figure}[t]
 \centering
 \noindent\includegraphics[width=2in]{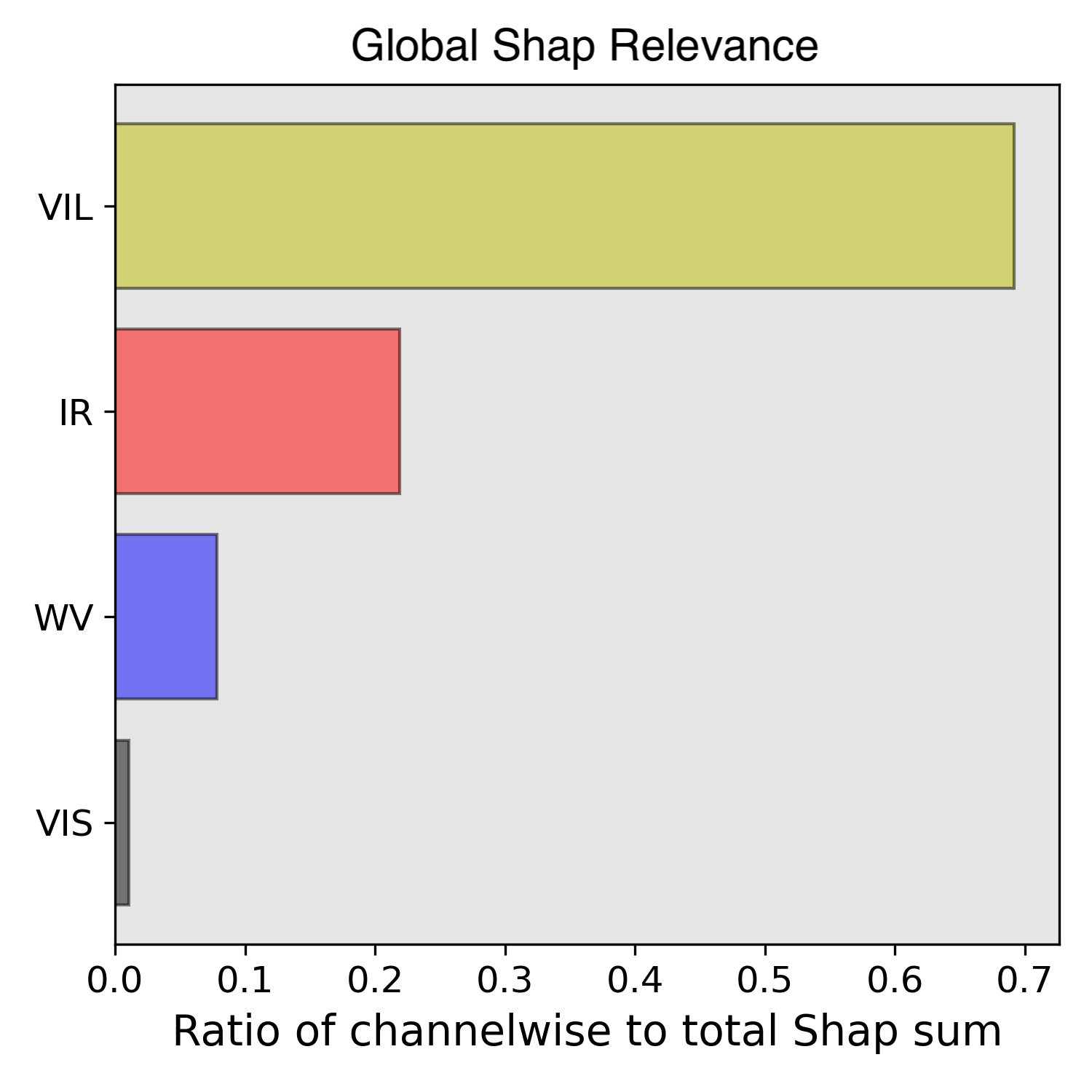}\\
 \caption{Global SHAP ratio on the validation dataset. The channel-wise ratio (i.e., sum across each input variables) of all SHAP values to the total SHAP sum. These SHAP values were evaluated on the entire validation dataset.}\label{shap_global}
\end{figure}

The additive property of SHAP values can be extended beyond this \textit{local} (i.e., one sample or case) explanation. The SHAP values can be summed across all dataset examples to get a similar \textit{global} explanation to what permutation importance gave us. The channel-wise sum across all examples in the validation dataset are shown in Fig. \ref{shap_global}. The result is the same as permutation importance, showing that the vertically integrated liquid is most important, followed by the infrared brightness temperature, water vapor brightness temperature and then visible reflectance. It is encouraging to get the same result from two different XAI methods, which builds confidence in the end result. 

While the SHAP discussion has been centered on the classification task, the same analysis can be done on the regression task but is not done here for brevity. Similarly, the discussion in this paper has been focused on neural networks, but SHAP can also be applied to the traditional methods of Part 1. For more examples of SHAP being used in the meteorological research readers can look over these references: \citet{Gensini2021}, \citet{Flora2022a}, \citet{Flora2022b}, \citet{Griffin2022}, \citet{Mamalakis2022}, \citet{Mamalakis2022b} and \citet{vanStraaten2022}.

\section{Summary \label{subsec:Summary}}
This manuscript is the second of a pair of machine learning tutorial papers designed for the operational meteorological community. The main focus of this paper was the plain language discussion of neural networks. More specifically the neural networks discussed included artificial neural networks (i.e., multi-layer perceptrons; ANN), convolutional neural networks (CNN) and "U" shaped networks (U-Net). Similar to Part 1 of this tutorial series \citep[i.e.,][]{Chase2022}, the goal of this paper was to provide an overview of the many terms involved in neural networks while also providing entry level intuition of each method and their training procedures. Furthermore, the same simple meteorological example using the Storm EVent ImageRy dataset \citep[SEVIR; ][]{Veillett2020} to identify lightning presence and amount was reconducted with the neural network methods to allow for direct comparison of all machine learning methods discussed in both parts of the series. Explicitly summarizing the results of this paper we: 
\begin{enumerate}
    \item{Discussed the various nuances and terms associated with neural networks (Section 2)}
    \item{Discussed three different neural network architectures in detail (Section 2.a)}
    \item{Demonstrated a classification and regression task to diagnose the presence and number of lightning flashes in a satellite image. (Section 3.de)}
    \item{Showed two eXplainable Artificial Intelligence techniques applied to a CNN (Section Section 3.de)}
    \item{Released python code to conduct all steps and examples in this manuscript (see \textit{Data Availability Statement})}
\end{enumerate}

As technology continually advances, unprecedented meteorological measurements and simulations will continue to occur. For example, the GOES-R series of geostationary satellite provides 0.5 km grid spacing of visible imagery that was only previously obtainable from polar orbiting satellites (e.g., MODIS). Another example includes the growing efforts to begin global simulations of weather using convective allowing horizontal grid spacing \citep[e.g., < 4 km;][]{Stevens2019}. With these improved measurements and simulations come daunting increases of dataset sizes and then potentially information overload (i.e., too much data to use). Thus, it is imperative that meteorologists are familiar with tools that can reduce their individual burden. Machine learning is poised to handle the future terabytes/petabytes of meteorological data and potentially can provide valuable tools for meteorologists to make trustworthy and well informed data-driven decisions.

\clearpage
\acknowledgments
This material is based upon work supported by the National Science Foundation under Grant No. ICER-2019758, supporting authors RJC, and AM. Author DRH was provided support by NOAA/Office of Oceanic and Atmospheric Research under NOAA-University of Oklahoma Cooperative Agreement number NA21OAR4320204, U.S. Department of Commerce. The scientific results and conclusions, as well as any views or opinions expressed herein, are those of the authors and do not necessarily reflect the views of NOAA or the Department of Commerce.

We want to acknowledge the work put forth by the authors of the SEVIR dataset (Mark S. Veillette, Siddharth Samsi and Christopher J. Mattioli) for making a high-quality free dataset. We would also like to acknowledge the open-source python community for providing their tools for free. Specifically, we acknowledge Google Colab \citep{Bisong2019}, Anaconda \citep{anaconda}, scikit-learn \citep{scikit-learn}, Pandas \citep{mckinney-proc-scipy-2010}, Numpy \citep{harris2020array} and Jupyter \citep{Kluyver2016jupyter}. 


%
%
\datastatement
As an effort to accelerate the use and trust of machine learning within meteorology we have supplied a github repository with a code tutorial of a lot of the same things discussed in this paper. The latest version of github repository can be located here: \url{https://github.com/ai2es/WAF_ML_Tutorial_Part2}. If you are interested in the version of the repository that was available at time of publication please see the zendo archive of version 1 here: \url{https://zenodo.org/record/7011372}. The original github repo for SEVIR is located here: \url{https://github.com/MIT-AI-Accelerator/neurips-2020-sevir}. 


%

\appendix

\appendixtitle{Hyperparameter Tuning Specifics}
All the models shown in the paper are the result of a fairly extensive hyperparameter search. Each of the following figures contains the different hyperparameters that were varied. Note that only 100 models were trained for each model type (e.g., ANN regression was one model), so it is very possible that not all possible hyperparameter solution sets were run. The following figures are for the ANN, CNN and U-Net respectively and red indicates the best configuration choice for regression, blue indicates the best configuration choice for classification and purple means the best configuration choice for both model types. The best configurations were determined by the best performance on the validation dataset. 

\begin{figure}
\centerline{\includegraphics[width=3in]{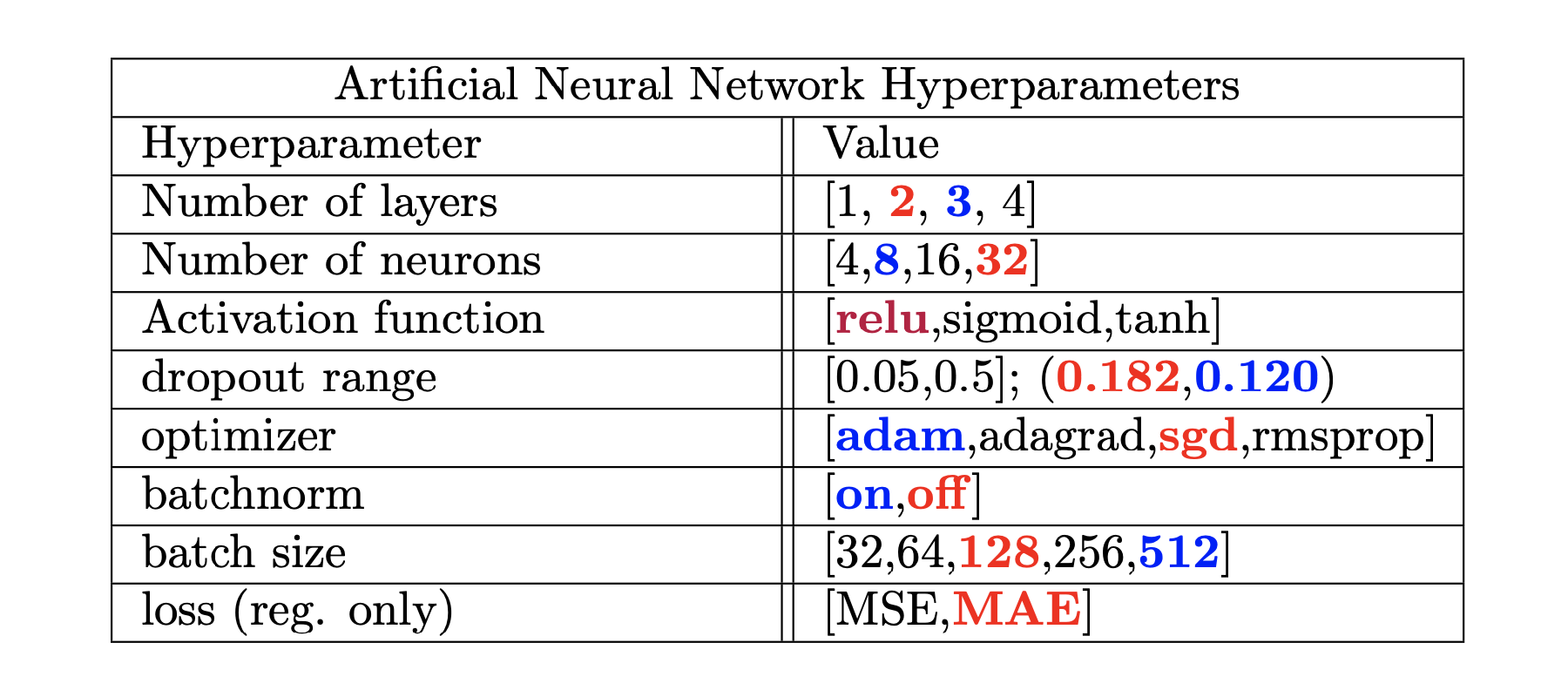}}
\caption{Figure showing the hyperparameters for the artificial neural networks}
\end{figure}

\begin{figure}
\centerline{\includegraphics[width=3in]{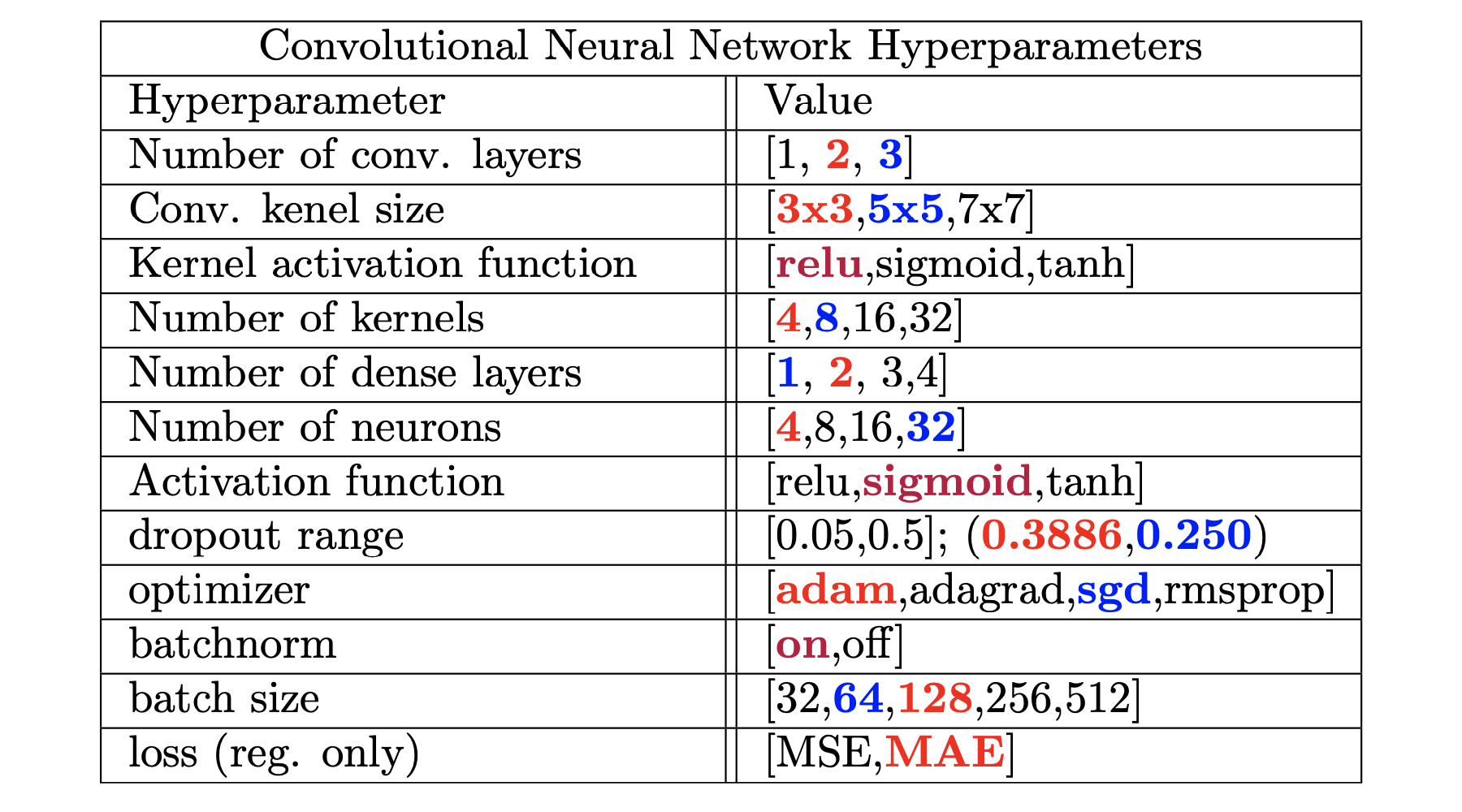}}
\caption{Figure showing the hyperparameters for the convolutional neural networks}
\end{figure}

\begin{figure}
\centerline{\includegraphics[width=3in]{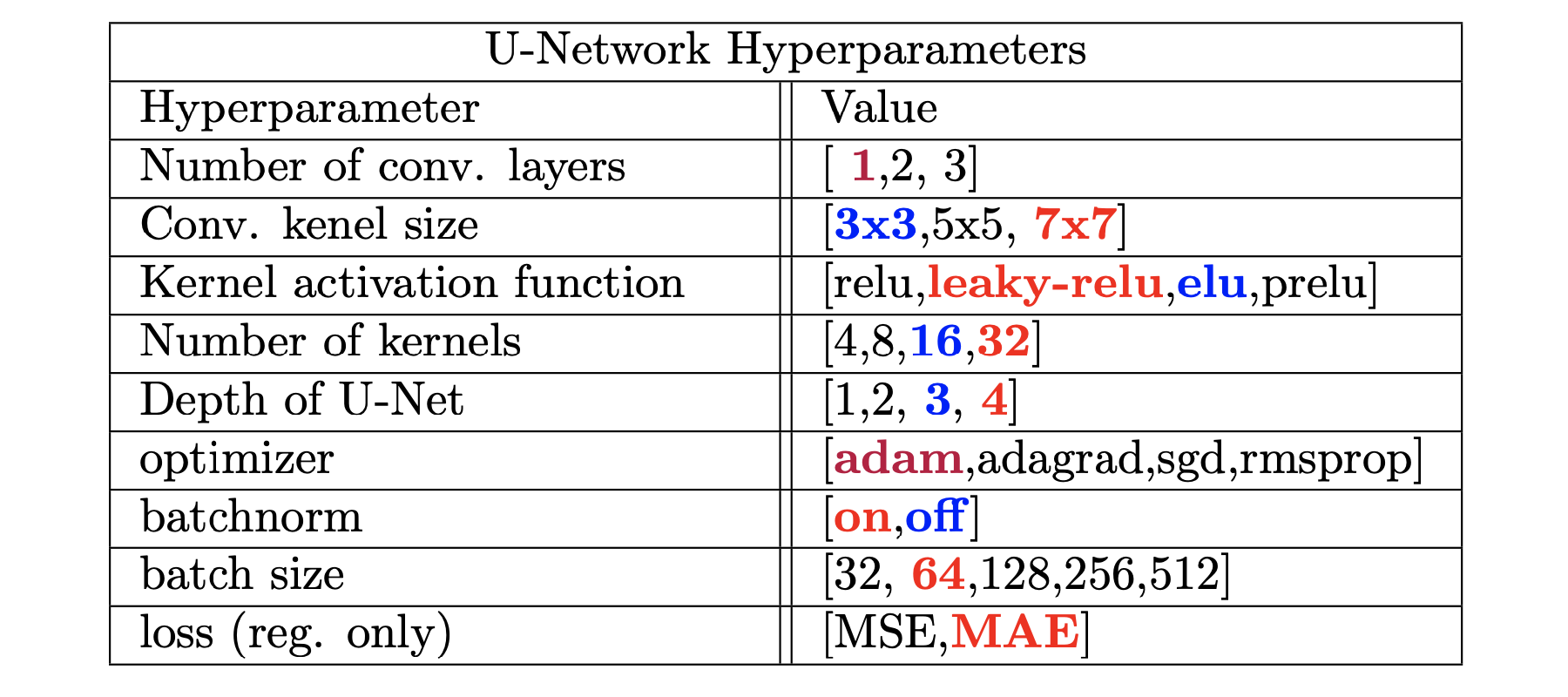}}
\caption{Figure showing the hyperparameters for the U-Network}
\end{figure}
\clearpage




%



\bibliographystyle{ametsocV6}
\bibliography{clean_references_part2}

\end{document}